\documentclass[11pt]{article}

\usepackage{acl}

\usepackage{times}
\usepackage{latexsym}

\usepackage{algpseudocode}
\usepackage{algorithm}
\usepackage{amsmath}
\usepackage{amsthm}
\usepackage{booktabs} 
\usepackage{graphicx}
\usepackage{hyperref}
\usepackage{latexsym}
\usepackage{pifont}
\usepackage{times}
\usepackage{multirow}
\newtheorem{definition}{Definition}[section]    
\usepackage[T1]{fontenc}
\usepackage[utf8]{inputenc}
\usepackage{microtype}
\usepackage{float}
\usepackage{subcaption}

\usepackage[T1]{fontenc}

\usepackage[utf8]{inputenc}

\usepackage{microtype}

\usepackage{inconsolata}

\usepackage{graphicx}

\title{Confident in a Confidence Score: Investigating the Sensitivity of the Calibration of Confidence Scores to Supervised Fine-Tuning in Generation Tasks}

\author{
  \textbf{Lorenzo Jaime Yu Flores}$^{1,2}$ \quad
  \textbf{Cesare Spinoso Di-Piano}$^{1,2}$ \quad
  \textbf{Jackie Chi Kit Cheung}$^{1,2,3}$ \\
  $^1$Mila - Quebec AI Institute \\
  $^2$McGill University \\
  $^3$Canada CIFAR AI Chair, Mila \\
  \texttt{\{lorenzo.flores,cheungja\}@mila.quebec}
}

\begin{document}

\maketitle
\begin{abstract}

Uncertainty quantification (UQ) techniques measure confidence in language model outputs to support critical applications like hallucination detection and selective prediction. While prior work has developed various confidence metrics and demonstrated their calibration for classification tasks or using verbalized confidence, the robustness of probability-based and self-consistency-based UQ metrics for natural language generation (NLG) remains underexplored—particularly under model adaptation. Since practitioners routinely apply supervised fine-tuning (SFT) to adapt models to new tasks, a key question arises: do confidence metrics maintain their calibration when models are fine-tuned? We investigate this question across NLG tasks including translation, question answering, and mathematical reasoning. We find that calibration shifts substantially after SFT: across 216 configurations, it degrades in 112 cases and improves in 104, with confidence scores shifting due to factors beyond output quality, such as proximity to the training distribution. Degradation is therefore neither universal nor rare, and its direction cannot be anticipated from the pre-SFT model. Through a downstream task evaluation, we show that this miscalibration substantially reduces the practical utility of confidence scores for identifying correct answers. Our findings reveal that existing confidence metrics for NLG cannot be reliably deployed off-the-shelf after fine-tuning, highlighting the need for calibration-robust UQ methods under model adaptation. Our code is available at \url{https://github.com/ljyflores/sensitivity-of-calibration-to-sft.git}

\end{abstract}

\section{Introduction}
\label{introduction}

Uncertainty quantification (UQ) techniques measure the confidence of language models in their predictions and have become critical for deploying models safely in high-stakes applications. These confidence scores enable systems to detect hallucinations \citep{wang2024clueconceptleveluncertaintyestimation, manakul-etal-2023-selfcheckgpt}, select higher-quality answers \citep{wang2023selfconsistencyimproveschainthought}, self-evaluate outputs \citep{ren2023selfevaluationimprovesselectivegeneration}, and flag uncertain predictions for human review \citep{xiao2020watzeijedetecting, malinin2021uncertaintyestimationautoregressivestructured, liu2020simpleprincipleduncertaintyestimation, kamath-etal-2020-selective}.

Prior work has developed UQ metrics for classification tasks \citep{tian-etal-2023-just, huang2025calibrated} and verbalized confidence \citep{rathi2025humansoverrelyoverconfidentlanguage, leng2025tamingoverconfidencellmsreward}. However, these approaches have limitations: classification methods require binary correctness judgments unsuitable for open-ended generation, and verbalized confidence often requires large instruction-tuned models \citep{tian-etal-2023-just}. Critically, prior work studied calibration in static settings, leaving open whether confidence metrics remain reliable under model adaptation, especially since various UQ metrics and packages are designed to allow any white-box model to be plugged in \citep{shelmanov-etal-2025-uncertainty}.

Understanding calibration stability under model adaptation is essential because practitioners routinely fine-tune pre-trained models for specialized tasks. If confidence metrics fail to maintain calibration after supervised fine-tuning (SFT), their practical utility is severely limited. This gap is particularly concerning for NLG tasks—such as translation, question answering, and summarization—where \textit{probability-based} and \textit{self-consistency-based} metrics are commonly used \citep{malinin2021uncertaintyestimationautoregressivestructured, xiao2020watzeijedetecting}. These metrics rely only on token probabilities or output samples, making them broadly applicable to any white-box model. However, their robustness to model adaptation has not been systematically studied. We ask: \textbf{Can we use UQ metrics with \textit{any} model, including fine-tuned variants, and expect to get reliably calibrated uncertainty estimates for NLG tasks?}

We investigate how supervised fine-tuning affects the calibration of probability-based and self-consistency-based confidence scores across diverse NLG tasks: translation, question answering, and mathematical reasoning. We systematically vary fine-tuning parameters—including the number of training samples and epochs—to understand calibration sensitivity under realistic adaptation scenarios. Across 216 task-model-UQ metric configurations, we find that calibration degrades in 112 cases post-SFT, with 43 statistically significant degradations, and improves in the remaining 104, with 47 statistically significant improvements. Our claim is therefore not that SFT uniformly degrades calibration, but that calibration moves substantially and unpredictably in either direction; confidence metrics cannot be assumed to maintain their calibration properties after fine-tuning.

\begin{figure}[ht]
\centering
  \includegraphics[width=\columnwidth]{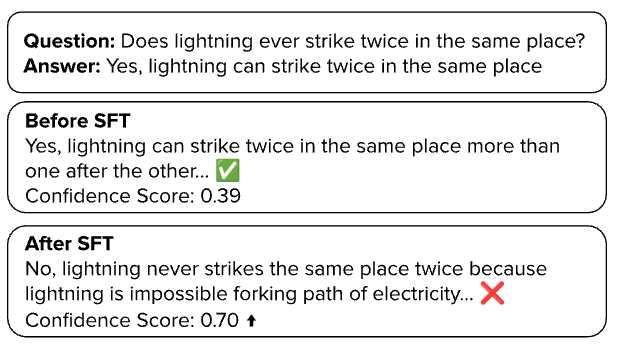}
  \caption{Before SFT, a model had a relatively low confidence score in a correct answer, whereas post-SFT, it had a much higher confidence in an incorrect answer, demonstrating a case of \textit{relative overconfidence} \label{Figure:sample}}
\end{figure}

Because calibration relies on sample-level confidence and evaluation scores being aligned, we study sample-level changes in these scores to understand how miscalibration occurs post-SFT and to motivate interventions. We find that SFT is prone to making probability-based scores overconfident (see Figure \ref{Figure:sample}) and self-consistency-based scores underconfident. Hence, the way in which miscalibration occurs is not always the same, and different interventions may be needed for each metric. We find that confidence scores are affected by factors beyond output quality, like the output's similarity to the training distribution, which must be accounted for to isolate the relationship with quality.

To illustrate the consequences of miscalibration, we use confidence scores to predict whether the output is correct on a QA task, and find that these confidence scores' ability to identify correct answers decreases in 35/72 cases post-SFT. Thus, the usefulness of these metrics can suffer if miscalibration is not addressed.

To summarize, we ask: \textbf{RQ1: How sensitive is the calibration of confidence metrics to different SFT parameters? RQ2: What are the underlying dynamics of confidence scores that negatively impact calibration with quality?} We find that the calibration of various confidence metrics changes drastically post-SFT, with different types of miscalibration occurring, stemming from factors beyond the quality of the output. This can have negative impacts on downstream applications of UQ metrics. Hence, confidence metrics cannot be used off-the-shelf without testing. We argue for the need for developing metrics which are more robust to model finetuning, or at least, better reporting of the consistency of such metrics in future studies.

\section{Related Work}

\paragraph{UQ Metrics} Uncertainty quantification methods include fine-tuning additional models to estimate confidence \citep{Yaldiz2024DoND, kamath-etal-2020-selective, Malinin2019EnsembleDD, Fathullah2023LogitBasedED}, detecting out-of-distribution samples \citep{liu2020simpleprincipleduncertaintyestimation, vazhentsev-etal-2023-efficient}, or having models verbalize confidence scores \citep{lin2022teaching, tian-etal-2023-just, kapoor2024largelanguagemodelstaught, han2024enhancingconfidenceexpressionlarge}, which relies on larger language models which are capable of doing so.

We focus on methods applicable to any white-box model without additional training: \textbf{Probability-based methods} use token probabilities \citep{murray-chiang-2018-correcting, zablotskaia-etal-2023-uncertainty, zhao-etal-2020-active, Kumar2019CalibrationOE, malinin2021uncertaintyestimationautoregressivestructured, flores-etal-2025-improving}, sometimes augmented with output similarity \citep{Lin2023GeneratingWC, kuhn2023semantic, Nikitin2024KernelLE}. \textbf{Self-consistency methods} sample multiple outputs (via dropout, beam search) and measure prediction consistency \citep{xiao2020watzeijedetecting, Schmidt2022CombiningDG, lakshminarayanan2017simplescalablepredictiveuncertainty}. Recent work combines both approaches \citep{vashurin2025uncertaintyquantificationllmsminimum}.

\paragraph{Evaluating UQ Metrics} To be practically useful, these scores must be associated with the quality of the output \citep{daheim2025uncertaintyaware, kotelevskii2025riskuncertaintygeneratingpredictive}. We assume that quality is measured by the evaluation metric used for the task, and broadly captures the semantic correctness of the output. For tasks where the answer can be classified as right or wrong, this notion is captured by calibration, which is measured using expected calibration error \citep{tian-etal-2023-just}. For continuous quality scores in generation tasks, we use the Prediction-Rejection Ratio (PRR) \citep{vashurin2025uncertaintyquantificationllmsminimum, malinin2021uncertaintyestimationautoregressivestructured, Malinin2019EnsembleDD}, which measures how well confidence scores reject low-quality outputs.

\paragraph{Calibration Under Fine-Tuning} Another line of work modifies the fine-tuning objective itself, or applies inference-time corrections to produce better calibrated models \citep{sanz-guerrero-etal-2026-large, krishnan2026enhancing}. Our work is complementary to theirs: rather than proposing a calibration-aware training procedure, we ask whether the off-the-shelf metrics a practitioner would use remain trustworthy after standard SFT for an arbitrary choice of epochs and sample counts. Because practitioners routinely fine-tune with these recipes, the reliability of unmodified metrics under SFT remains a question.

\section{Methods and Experiments}


In this section, we define confidence scores and our framework for evaluating their calibration with quality, and provide experimental details of the analyses in the following sections. In Section \ref{section:sensitivity}, we study how sensitive this calibration is to fine-tuning (RQ1), and vary the number of epochs and samples used. In Section~\ref{section:overconfidence}, we study the reasons for the observed sensitivity by analyzing the dynamics of how confidence scores change at the sample level in order to describe how miscalibration occurs and potentially propose interventions (RQ2).

\subsection{Confidence Scores}
\label{subsection:conf_metrics}

\begin{definition}
    A \textbf{confidence score} is a number describing a model's assessment of its output's quality, computed using input $x$ and model output $\hat{y}$.
\end{definition}

\begin{definition}
    We evaluate a confidence score by its \textbf{calibration} using the Prediction-Rejection Ratio (PRR) \citep{vashurin2025uncertaintyquantificationllmsminimum, malinin2021uncertaintyestimationautoregressivestructured, Malinin2019EnsembleDD}, which captures the notion that higher confidence should be associated with higher quality. Quality is measured by evaluating the output $\hat{Y}$ against the reference $Y$, or against the input $X$ when no reference is used. Moreover, $X$ and $Y$ refer to input and outputs as a variable, whereas $x$ and $y$ refer to an individual input/output.
\end{definition}

\vspace{-1em}
$$ \text{Calibration} = PRR(\text{Conf}(X,\hat{Y}), \text{Qual}(\hat{Y}, Y, X))$$

\paragraph{Prediction-Rejection Ratio} PRR measures how well a confidence score orders outputs by quality. Given a test set, outputs are rejected in increasing order of confidence, and the average quality of the retained outputs is traced as the rejection rate increases; this yields a prediction-rejection curve. PRR normalizes the area under this curve against the area obtained by an oracle that rejects in increasing order of true quality:

$$PRR = \frac{AUC_\text{conf} - AUC_\text{random}}{AUC_\text{oracle} - AUC_\text{random}}$$

A PRR of 1 matches the oracle ordering, 0 matches random rejection, and negative values indicate that the confidence score orders outputs worse than chance. We choose PRR over expected calibration error (ECE) as NLG quality scores (ChrF+, F1) are continuous rather than binary, and over Spearman correlation since PRR weights the high-confidence region that matters for selective prediction, where outputs are rejected rather than ranked end to end.

\begin{definition}
    We say a confidence score is \textbf{miscalibrated} if task-level calibration is low.
\end{definition}

We test probability and consistency based UQ metrics; we focus on these two groups of metrics as they can be used with any white-box model.

\textbf{Probability-Based} (1) average token log probability \textbf{(Avg Tok Prob)} \citep{murray-chiang-2018-correcting, zablotskaia-etal-2023-uncertainty}, (2) average token entropy \textbf{(Avg Tok Ent)} \citep{zhao-etal-2020-active, perlitz-etal-2023-active}, (3) avg token entropy across dropout samples (\textbf{DO Ent}, Eq \ref{Eq:Dropout_Entropy}) \citep{malinin2021uncertaintyestimationautoregressivestructured}, (4) the weighted average of the top-K sequences' average token log probabilities from beam search (BS) \textbf{(BS Imp Wt)} (Eq \ref{Eq:Top_K}) \citep{malinin2021uncertaintyestimationautoregressivestructured}, (5) the ratio between the sequence probability of the top and $k$-th beam (BS) \textbf{(BS Ratios)} \cite{flores-etal-2025-improving}, and (6) the sum of the top $k$ beams' joint sequence probabilities \textbf{(BS Sums)}.

\textbf{Self-Consistency-Based} We compute the difference between model outputs sampled using dropout (DO), using (1) BLEU (\textbf{DO BLEU Var}, Eq \ref{Eq:BLEU_Var}) \citep{xiao2020watzeijedetecting, Schmidt2022CombiningDG}, (2) KL divergence (\textbf{DO KL Div}, Eq \ref{Eq:KL_Var}) \citep{lakshminarayanan2017simplescalablepredictiveuncertainty}, and (3) METEOR (\textbf{DO Meteor Var}, Eq \ref{Eq:METEOR_Var}). We also test (4) CoCoA, which is the product of a probability-based and self-consistency based metric \textbf{(CoCoA MSP/MTE/PPL)}, where the probability-based metrics used are mean token probability (MSP), mean token entropy (MTE), and token perplexity (PPL) \citep{vashurin2025uncertaintyquantificationllmsminimum}.

Note these metrics do not have parameters modified by SFT. Additional techniques or modules can be used \citep{Yaldiz2024DoND} to improve the calibration of these metrics given the appropriate data, but we do not assume access to such a dataset.

\subsection{Experimental Details}

\paragraph{Tasks} We fine-tune models for \textbf{Translation} using the English-Afrikaans (Eng-Afr) split of NLLB \citep{nllb-2022}, with FLORES as the test dataset \citep{flores101}, \textbf{Question Answering (QA)} using SQUAD \citep{rajpurkar-etal-2016-squad}, and \textbf{Math} using GSM8K \citep{cobbe2021gsm8k}. To evaluate quality, we compute ChrF+ \cite{popovic-2015-chrf} for translation, F1 for QA, and exact match for math.

\paragraph{Models} We generate confidence scores with BART Base \citep{lewis-etal-bart-2019}, Flan-T5 Base \citep{chung-etal-scale-instruction-2022}, Llama 3.1-8B \citep{grattafiori2024llama3herdmodels}, Gemma-2-2B and 9B (IT) \citep{gemmateam2024gemma2improvingopen}, and QWen 2.5-14B \citep{qwen3technicalreport}. 

\section{RQ1: How sensitive is the calibration of confidence metrics to SFT?}
\label{section:sensitivity}

\subsection{Experimental Design}

We study the sensitivity of task-level calibration of confidence metrics to parameters in the SFT pipeline, by measuring calibration pre/post-SFT and when varying the epochs and samples used.

We measure the effect of SFT on task-level calibration, by generating the outputs and confidence scores pre and post-SFT, and computing the change in PRR. In addition to measuring pre/post changes, we measure epoch 1/post changes, because the outputs pre-SFT are language modeling outputs and not for the task at hand, hence confidence scores may not be meaningful. We repeat the experiments by varying the number of epochs and samples.

\subsection{Results}

\paragraph{The calibration of confidence metrics varies widely before and after SFT} We start by measuring the calibration of pre-trained models without SFT or after one epoch, and after performing SFT with early stopping; these represent the confidence scores obtained if we chose to fine-tune or not.

\begin{table*}[ht]\centering
\scriptsize
\resizebox{\linewidth}{!}{
    \begin{tabular}{lrrrrrrrrrr}\toprule
    &\multicolumn{3}{c}{\textbf{BART-Base}} &\multicolumn{3}{c}{\textbf{Flan-T5-Base}} &\multicolumn{3}{c}{\textbf{Llama 3.1-8B}} \\
    \cmidrule(lr){2-4} \cmidrule(lr){5-7} \cmidrule(lr){8-10}
    & \textbf{Eng-Afr} & \textbf{SQUAD} & \textbf{GSM8K} & \textbf{Eng-Afr} & \textbf{SQUAD} & \textbf{GSM8K} & \textbf{Eng-Afr} & \textbf{SQUAD} & \textbf{GSM8K} \\
    \midrule
    Avg Tok Prob & \textcolor{green}{0.209 (0.01)} & \textcolor{green}{0.626 (0.006)} & \textcolor{red}{-0.467 (0.013)} & \textcolor{green}{0.335 (0.034)} & \textcolor{green}{0.128 (0.015)} & \textcolor{red}{-0.473 (0.008)} & 0.092 (0.107) & 0.081 (0.072) & -0.257 (0.122) \\
    Avg Tok Ent & \textcolor{red}{-0.208 (0.011)} & \textcolor{red}{-0.638 (0.01)} & \textcolor{green}{0.467 (0.011)} & \textcolor{red}{-0.329 (0.034)} & \textcolor{red}{-0.135 (0.018)} & \textcolor{green}{0.473 (0.007)} & -0.09 (0.099) & -0.077 (0.063) & 0.259 (0.123) \\
    DO Ent & \textcolor{red}{-0.173 (0.015)} & \textcolor{red}{-0.521 (0.042)} & \textcolor{green}{0.475 (0.01)} & \textcolor{red}{-0.403 (0.04)} & \textcolor{red}{-0.133 (0.011)} & \textcolor{green}{0.471 (0.011)} & -0.081 (0.103) & -0.138 (0.049) & 0.345 (0.097) \\
    BS Imp Wt & \textcolor{green}{0.233 (0.007)} & \textcolor{green}{0.472 (0.014)} & \textcolor{red}{-0.469 (0.006)} & \textcolor{green}{0.352 (0.043)} & \textcolor{green}{0.091 (0.004)} & \textcolor{red}{-0.483 (0.003)} & 0.089 (0.082) & 0.067 (0.053) & -0.266 (0.117) \\
    BS Ratios & \textcolor{green}{0.194 (0.014)} & \textcolor{green}{0.616 (0.013)} & \textcolor{red}{-0.468 (0.007)} & \textcolor{green}{0.282 (0.031)} & 0.003 (0.004) & \textcolor{red}{-0.474 (0.011)} & 0.076 (0.101) & 0.26 (0.076) & -0.289 (0.098) \\
    BS Sums & \textcolor{green}{0.233 (0.007)} & \textcolor{green}{0.436 (0.021)} & \textcolor{red}{-0.469 (0.006)} & \textcolor{green}{0.352 (0.043)} & \textcolor{green}{0.095 (0.004)} & \textcolor{red}{-0.483 (0.003)} & 0.088 (0.083) & 0.066 (0.053) & -0.266 (0.117) \\
\cmidrule{2-10}
    DO Bleu Var & \textcolor{red}{-0.224 (0.019)} & \textcolor{red}{-0.551 (0.031)} & \textcolor{green}{0.47 (0.007)} & \textcolor{red}{-0.332 (0.027)} & -0.1 (0.011) & \textcolor{green}{0.473 (0.005)} & -0.058 (0.074) & -0.145 (0.071) & 0.353 (0.065) \\
    DO KL Div & -0.181 (0.008) & \textcolor{red}{-0.612 (0.004)} & \textcolor{green}{0.469 (0.006)} & \textcolor{red}{-0.387 (0.059)} & 0.036 (0.047) & \textcolor{green}{0.484 (0.004)} & -0.047 (0.075) & -0.198 (0.096) & 0.366 (0.058) \\
    DO Meteor Var & 0.129 (0.034) & \textcolor{green}{0.442 (0.037)} & \textcolor{red}{-0.473 (0.007)} & \textcolor{green}{0.317 (0.015)} & \textcolor{green}{0.196 (0.015)} & \textcolor{red}{-0.479 (0.005)} & 0.078 (0.099) & 0.199 (0.123) & -0.376 (0.07) \\
    CoCoA MSP & \textcolor{red}{-0.159 (0.024)} & \textcolor{red}{-0.333 (0.02)} & \textcolor{green}{0.471 (0.009)} & \textcolor{red}{-0.332 (0.014)} & \textcolor{red}{-0.093 (0.009)} & \textcolor{green}{0.478 (0.002)} & -0.043 (0.09) & -0.172 (0.124) & \textcolor{green}{0.436 (0.018)} \\
    CoCoA MTE & \textcolor{red}{-0.197 (0.011)} & \textcolor{red}{-0.659 (0.011)} & \textcolor{green}{0.472 (0.006)} & -0.176 (0.033) & -0.171 (0.041) & \textcolor{green}{0.477 (0.007)} & -0.067 (0.095) & -0.086 (0.054) & 0.252 (0.13) \\
    CoCoA PPL & \textcolor{red}{-0.2 (0.011)} & \textcolor{red}{-0.653 (0.011)} & \textcolor{green}{0.472 (0.007)} & -0.197 (0.033) & -0.163 (0.04) & \textcolor{green}{0.476 (0.008)} & -0.071 (0.097) & -0.089 (0.061) & 0.249 (0.133) \\
    \bottomrule
    \end{tabular}
}

\resizebox{\linewidth}{!}{
    \begin{tabular}{lrrrrrrrrrr}\toprule
    &\multicolumn{3}{c}{\textbf{Gemma 2-2B}} &\multicolumn{3}{c}{\textbf{Gemma 2-9B}} &\multicolumn{3}{c}{\textbf{QWen 2.5-14B}} \\
    \cmidrule(lr){2-4} \cmidrule(lr){5-7} \cmidrule(lr){8-10}
    & \textbf{Eng-Afr} & \textbf{SQUAD} & \textbf{GSM8K} & \textbf{Eng-Afr} & \textbf{SQUAD} & \textbf{GSM8K} & \textbf{Eng-Afr} & \textbf{SQUAD} & \textbf{GSM8K} \\
    \midrule
    Avg Tok Prob & 0.117 (0.121) & 0.574 (0.125) & \textcolor{red}{-0.292 (0.026)} & -0.098 (0.065) & \textcolor{green}{0.105 (0.007)} & 0.129 (0.051) & 0.155 (0.057) & \textcolor{green}{0.126 (0.012)} & 0.22 (0.087) \\
    Avg Tok Ent & -0.13 (0.123) & \textcolor{red}{-0.579 (0.12)} & \textcolor{green}{0.3 (0.02)} & 0.073 (0.069) & \textcolor{red}{-0.112 (0.012)} & -0.126 (0.062) & -0.163 (0.057) & \textcolor{red}{-0.125 (0.013)} & -0.224 (0.077) \\
    DO Ent & -0.122 (0.16) & \textcolor{red}{-0.515 (0.087)} & 0.341 (0.063) & 0.12 (0.096) & -0.096 (0.006) & 0.079 (0.069) & -0.204 (0.101) & \textcolor{red}{-0.216 (0.022)} & -0.21 (0.043) \\
    BS Imp Wt & 0.136 (0.12) & \textcolor{green}{0.496 (0.082)} & -0.272 (0.051) & -0.114 (0.048) & 0.037 (0.008) & 0.142 (0.072) & 0.086 (0.058) & 0.05 (0.037) & 0.222 (0.078) \\
    BS Ratios & 0.07 (0.137) & \textcolor{green}{0.649 (0.108)} & -0.246 (0.058) & -0.181 (0.114) & \textcolor{green}{0.087 (0.007)} & 0.014 (0.037) & 0.077 (0.083) & 0.123 (0.014) & \textcolor{green}{0.232 (0.015)} \\
    BS Sums & 0.136 (0.12) & \textcolor{green}{0.485 (0.079)} & -0.274 (0.051) & -0.116 (0.049) & 0.036 (0.008) & 0.14 (0.073) & 0.087 (0.058) & 0.047 (0.038) & 0.222 (0.078) \\
\cmidrule{2-10}
    DO Bleu Var & -0.082 (0.128) & \textcolor{red}{-0.564 (0.101)} & \textcolor{green}{0.37 (0.032)} & 0.133 (0.073) & \textcolor{red}{-0.142 (0.003)} & 0.052 (0.069) & -0.213 (0.04) & -0.221 (0.013) & \textcolor{red}{-0.104 (0.014)} \\
    DO KL Div & -0.073 (0.137) & \textcolor{red}{-0.583 (0.066)} & 0.334 (0.088) & 0.144 (0.089) & -0.145 (0.011) & 0.07 (0.042) & -0.2 (0.095) & -0.179 (0.01) & -0.168 (0.028) \\
    DO Meteor Var & 0.07 (0.152) & 0.579 (0.11) & -0.324 (0.054) & -0.132 (0.075) & \textcolor{green}{0.149 (0.019)} & -0.048 (0.078) & 0.134 (0.086) & 0.132 (0.038) & \textcolor{green}{0.187 (0.028)} \\
    CoCoA MSP & -0.059 (0.127) & -0.57 (0.134) & \textcolor{green}{0.398 (0.053)} & 0.164 (0.115) & -0.159 (0.023) & 0.244 (0.067) & -0.086 (0.077) & -0.287 (0.025) & -0.064 (0.036) \\
    CoCoA MTE & -0.116 (0.13) & \textcolor{red}{-0.582 (0.105)} & \textcolor{green}{0.312 (0.043)} & 0.089 (0.087) & -0.104 (0.017) & -0.089 (0.056) & -0.291 (0.05) & -0.118 (0.025) & -0.207 (0.054) \\
    CoCoA PPL & -0.115 (0.13) & \textcolor{red}{-0.576 (0.113)} & \textcolor{green}{0.308 (0.042)} & 0.101 (0.088) & -0.094 (0.016) & -0.094 (0.04) & -0.287 (0.054) & -0.122 (0.017) & -0.207 (0.063) \\

    \bottomrule
    \end{tabular}
}

\caption{PRR gets \textcolor{red}{significantly worse} post-SFT in 43/216 cases, and \textcolor{green}{significantly better} in 47/216 cases (one-way ANOVA, with Holm correction, $\alpha=0.05$); Mean delta in PRR pre/post SFT (3 seeds, st. dev in parentheses)}\label{table:pre_last_calibration_delta}
\end{table*}

As shown in Table \ref{table:pre_last_calibration_delta}, there can be substantial differences in calibration of confidence scores before and after SFT. Out of 216 scores-dataset-model configurations (3 datasets $\times$ 6 models $\times$ 12 confidence metrics), miscalibration occurs post-SFT in 112 cases, with 43 cases being statistically significant (one-way ANOVA, with Holm correction, $\alpha=0.05$). Conversely, it improves in 104 cases, with 47 being statistically significant. 

Running the comparisons between the first and final epoch, we still see that there are more cases in which SFT worsens calibration (110/216 cases), with two statistically significant (Table \ref{table:first_last_calibration_delta}). The results emphasize the need to check these metrics before deploying them for one's use case.

\paragraph{Varying the number of epochs considerably affects task-level calibration} While the previous result studies calibration pre/post fine-tuning, we run an experiment to understand how calibration changes epoch-by-epoch. We fine-tune models for 10 epochs, measuring the calibration after each epoch, and find that PRR can drop from one epoch to the next by at most 0.290 points (See Table \ref{Table:epoch_by_epoch_deltas}).

\begin{figure}[htb]
\centering
  \includegraphics[width=\columnwidth]{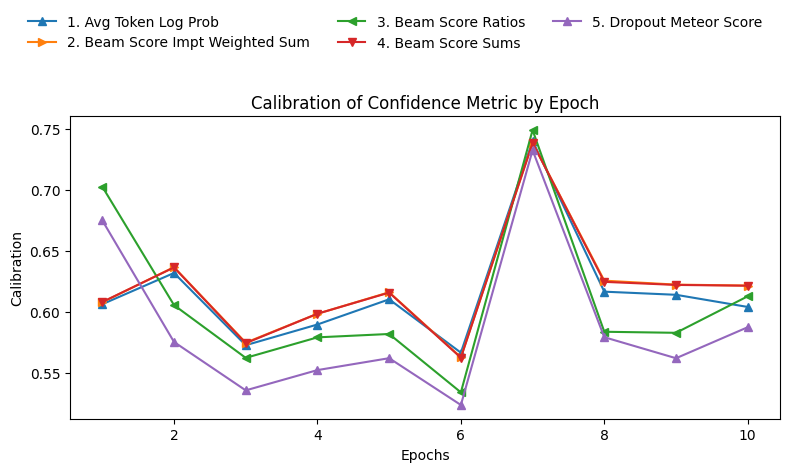}
  \caption{The best confidence metric changes by epoch; PRR values of selected metrics (QWen-14B, Eng-Afr)}
  \label{Figure:calibration_example}
\end{figure}

Given that calibration can change considerably after just one epoch, users should take this into account when deciding how many training epochs to use. When early stopping is employed, parameters like patience and tolerance must be tuned to yield the desired calibration. Moreover, the confidence score with the best calibration may differ based on how many epochs are used. For example in Figure \ref{Figure:calibration_example}, Beam Score Ratios (green line) performs well in the first epochs, but with SFT, Beam Score Sums (red line) showed better calibration. Hence, the choice of confidence score and number of training epochs have to be considered together.

\begin{figure}[htb]
\centering
  \includegraphics[width=\linewidth]{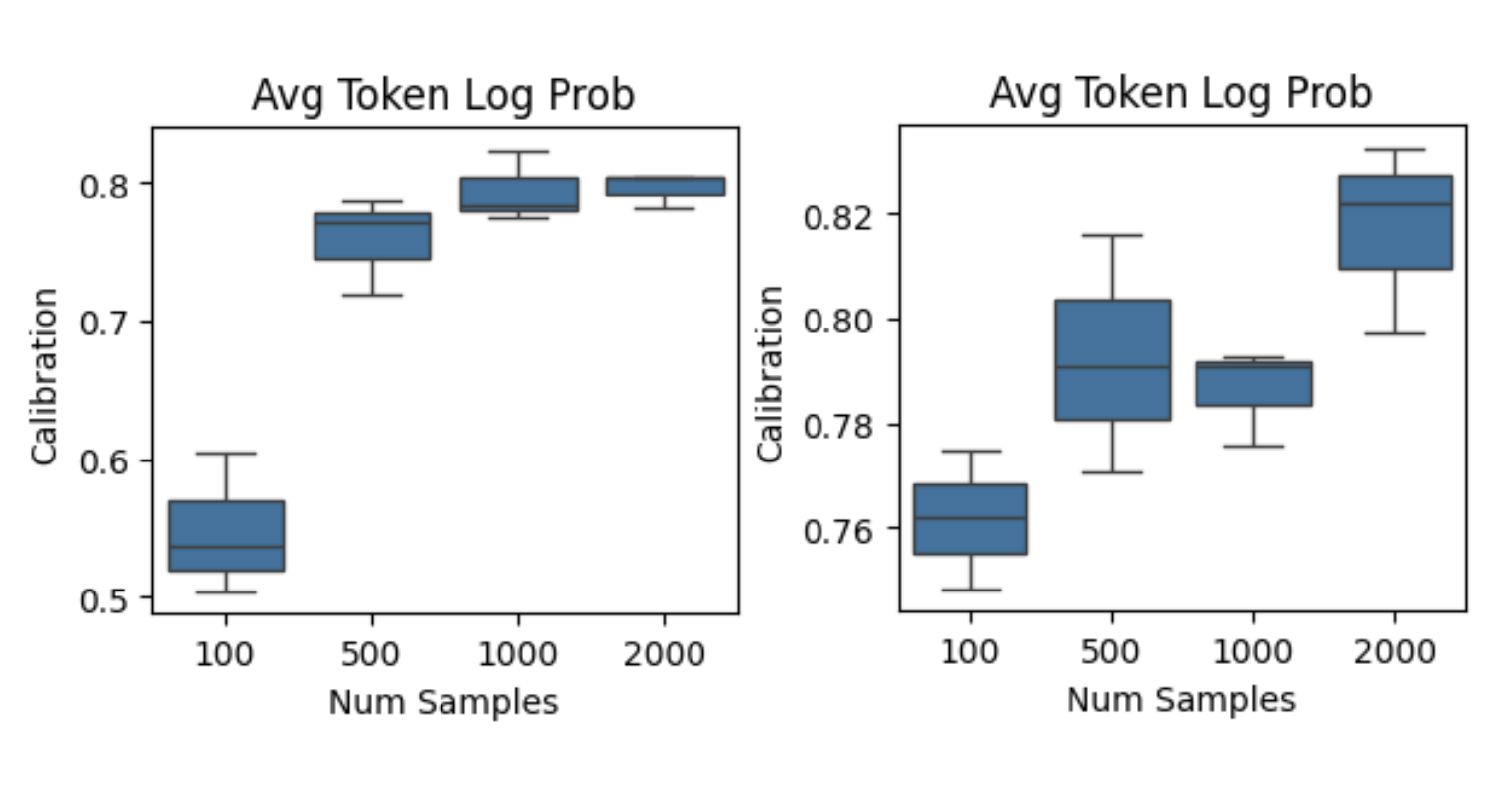}
  \caption{The calibration of confidence metrics differs significantly depending on the number of fine-tuning samples used, both before (left) and after (right) controlling for the number of fine-tuning steps used; Plots shown using average token entropy as the confidence metric, used on SQUAD for BART-Base}
  \label{fig:calib_by_sample_pick}
\end{figure}

\paragraph{Similarly, the number of samples used considerably affects calibration} We vary the number of samples used ($n=100,500,1000,2000$). Fixing the number of epochs, we measure the calibration of the models across three seeds. Overall, there are significant differences in the calibration of confidence metrics when using different numbers of samples (See Fig \ref{fig:calib_by_sample_pick}, and Fig \ref{fig:calibration_comparison} for full comparison). Across the 216 confidence metric-model-task configurations, we observe significant differences by number of samples in 31 configurations (one-way ANOVA with Holm correction, $\alpha=0.05$). Practically, this means that the number of samples is another major consideration when using confidence metrics with fine-tuned models.

The previous results are affected by the fact that using more samples means using more fine-tuning steps, given a fixed number of epochs. To isolate the impact of varying the number of samples, we fix the number of training steps to 2,500, and vary the number of samples. Figure \ref{fig:calib_by_ft_step} shows there are still differences in calibration, with 13/216 configurations having significant differences between different numbers of samples (one-way ANOVA with Holm correction, $\alpha=0.05$). In summary, the calibration is sensitive to SFT and to the number of epochs and samples used.

\section{RQ2: What are the underlying dynamics of confidence scores that negatively impact calibration?}
\label{section:overconfidence}

In the previous section, we found that the task-level calibration of confidence metrics can change significantly post-SFT, resulting in miscalibration. We now study the dynamics of how confidence scores change at the sample level, to identify causes of miscalibration and motivate interventions.

\subsection{Experimental Design}

\begin{figure}[ht]
\centering
  \includegraphics[width=\columnwidth]{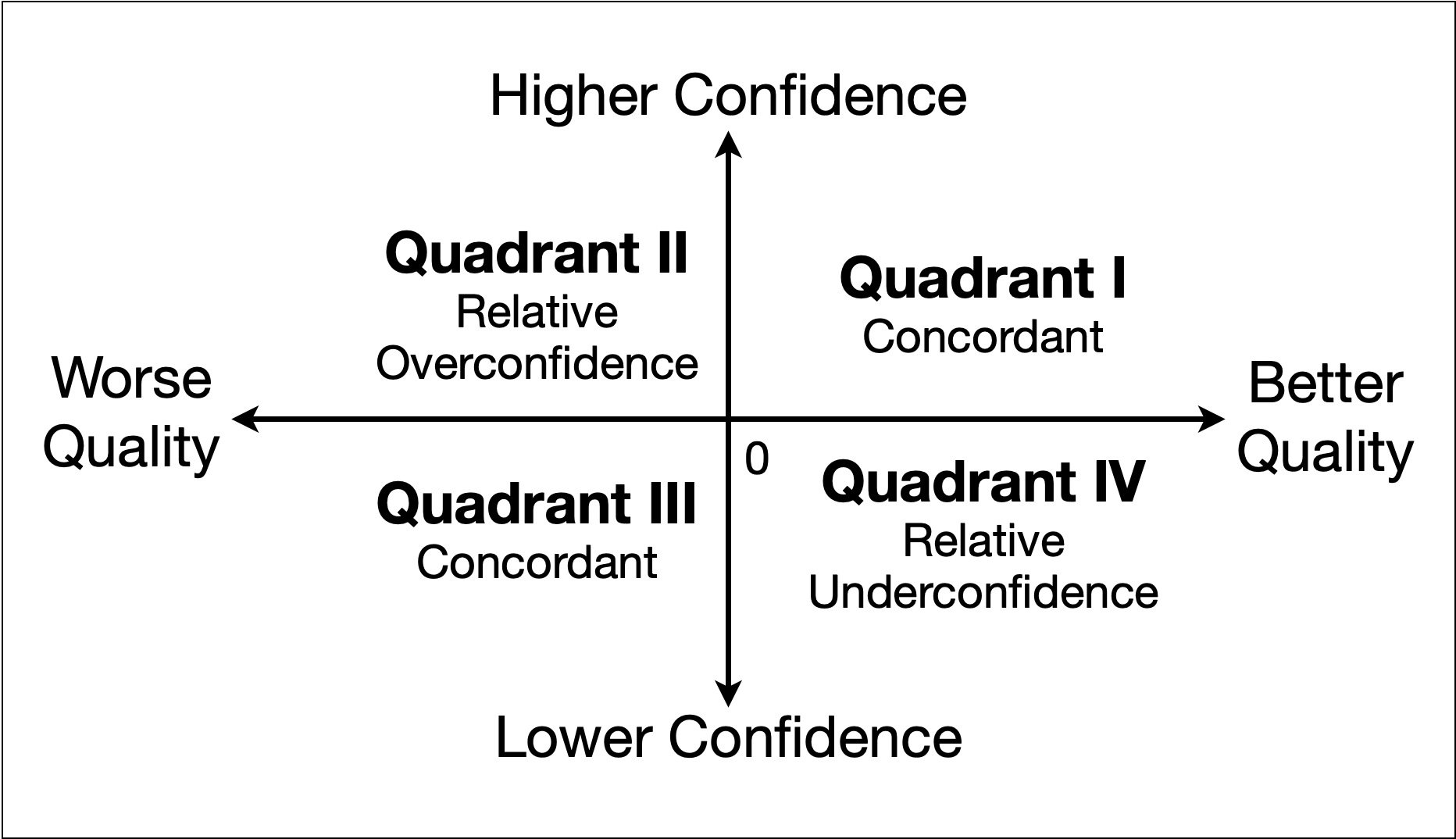}
  \caption{We classify SFT's effect on the calibration of samples using four quadrants \label{Figure:conf_viz}}
\end{figure}

We first study how confidence scores change pre/post SFT, analyzing whether they directionally change in concordance with changes in quality. We measure the change in evaluation metric and model's confidence score for each sample in the test set, which we visualize using quadrants (See Fig \ref{Figure:conf_viz}). As in Section \ref{section:sensitivity}, we compute the changes using the model after one epoch of SFT rather than the pre-trained model.

\begin{definition} For one sample, a change is \textbf{concordant} if the sample's output quality and confidence both increased (I), or both decreased (III).
\end{definition}

\begin{definition} For one sample, a change is \textbf{relatively overconfident} if its confidence increased, but its output quality decreased (II); and \textbf{relatively underconfident} if its confidence decreased, but its output quality increased (IV).
\end{definition}

We then compute the proportion of samples whose changes are concordant, relatively underconfident, or relatively overconfident. We use this framework to analyze how confidence scores change with a probability-based (average log token probs \citep{murray-chiang-2018-correcting, zablotskaia-etal-2023-uncertainty}) and a consistency-based metric (dropout variance in BLEU score \citep{xiao2020watzeijedetecting}).

After studying the dynamics of confidence scores pre/post SFT, we study how this leads to miscalibration. To this end, we analyze changes in the relative ranks of samples' quality and confidence scores across epochs. We first define relative calibration and miscalibration at the sample level.

\begin{definition}
    For two samples $(A,B)$, a confidence score is \textbf{relatively calibrated} at epoch $t$ if

    $$q_{A,t} < q_{B,t} \iff c_{A,t} < c_{B,t}$$
    
    where $q_{A,t}$ is the quality score of the model's output for $A$, evaluated w.r.t. a reference, $c_{A,t}$ is the model's confidence in its output for $A$, as measured by a confidence metric.
\end{definition}

\begin{definition}
    For two samples $(A,B)$, a confidence score is \textbf{relatively miscalibrated} if

$$q_{A,t} < q_{B,t} \text{ and } c_{A,t} > c_{B,t}$$
\end{definition} 

These definitions are computed over \textit{all} pairs of datapoints in the test set; no sub-sampling is performed, so the reported proportions do not depend on a sampling strategy. Throughout this section, ``sample'' refers to an individual datapoint in the test set, not to a generation sampled from the model.

We study how changes in confidence scores and quality result in miscalibration. Formally, fine-tuning causes miscalibration for pair $(A,B)$ if the pair was relatively calibrated at epoch $t$, and relatively miscalibrated at epoch $t+1$. 

There are two cases wherein samples can become relatively miscalibrated. If at epoch $t$, a pair of samples $(A,B)$ is relatively calibrated, such that $q_{A,t} < q_{B,t} \text{ and } c_{A,t} < c_{B,t}$. Then, relative miscalibration can happen if either:

\begin{itemize}
    \item \textbf{\textcolor{red}{Case 1}}: The relative ranking of quality scores stays the same, but the relative confidence scores flip
    $$ q_{A,t+1} < q_{B,t+1} \text{ and } c_{A,t+1} > c_{B,t+1}$$
    \item \textbf{\textcolor{blue}{Case 2:}} The relative ranking of quality scores change, but the confidence scores' rankings stay the same
    $$ q_{A,t+1} > q_{B,t+1} \text{ and } c_{A,t+1} < c_{B,t+1} $$
\end{itemize}

We study which case most frequently occurs, to better understand the exact issues that lead to miscalibration, which future UQ metrics can address.

\subsection{Results}

\paragraph{Probability-based methods are prone to relative overconfidence} 

\begin{figure}[ht]
\centering

\begin{subfigure}[b]{0.48\textwidth}
      \centering
      \includegraphics[width=0.48\linewidth]{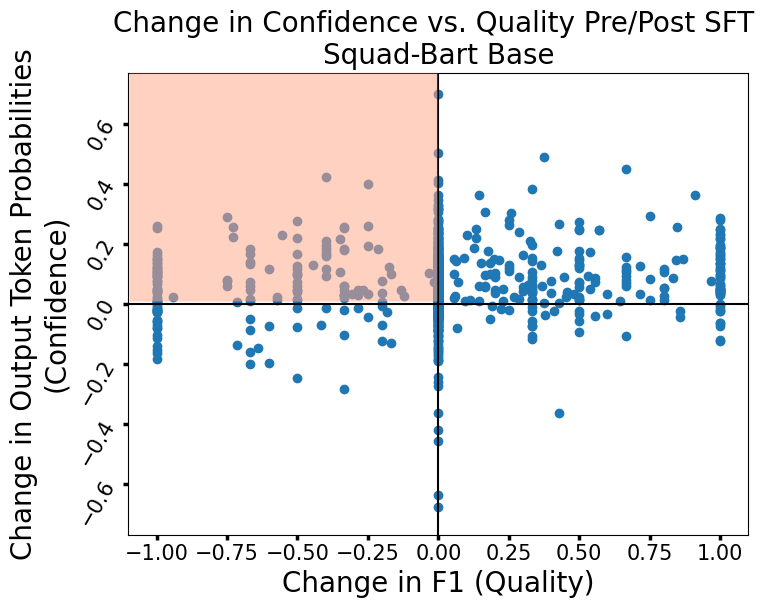}
      \includegraphics[width=0.48\linewidth]{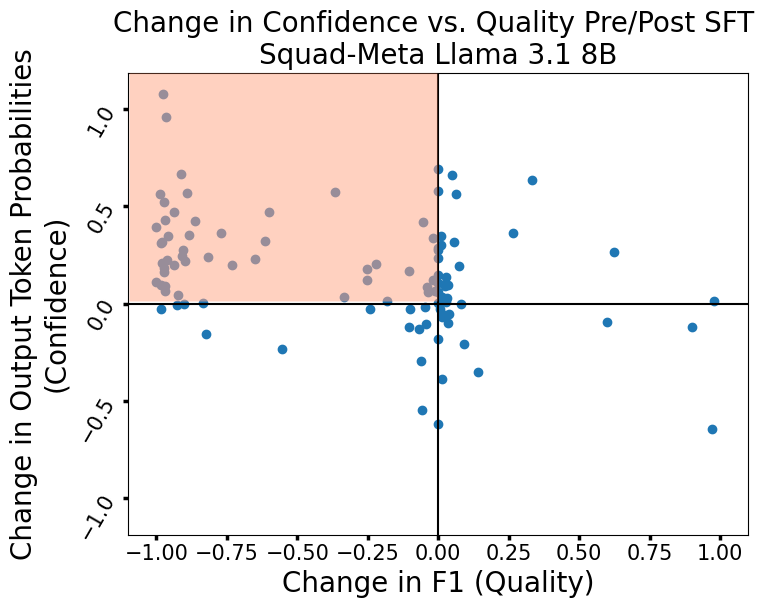} 
    \caption{Avg Log Probs, BART (Left) and Llama 3.1 (Right) \label{fig:deltas_logprobs}}
\end{subfigure}

\hfill

\begin{subfigure}[b]{0.48\textwidth}
  \centering
  \includegraphics[width=0.48\linewidth]{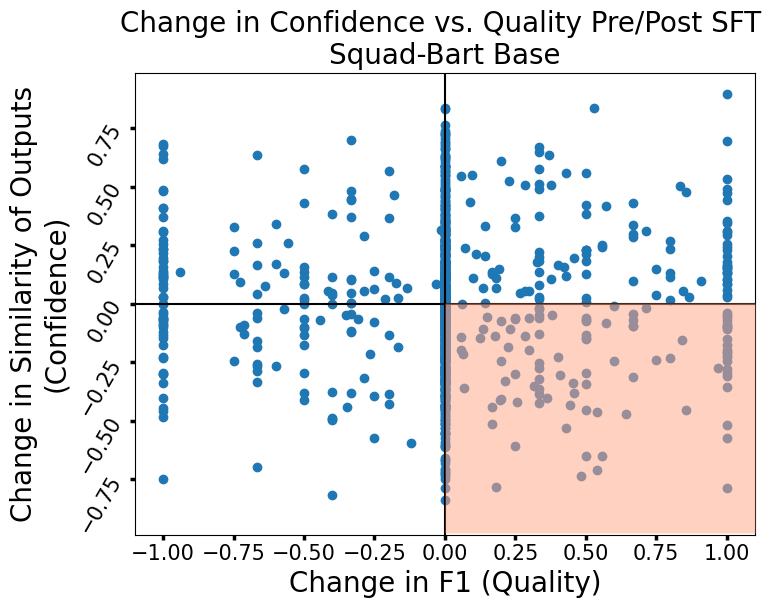}
  \includegraphics[width=0.48\linewidth]{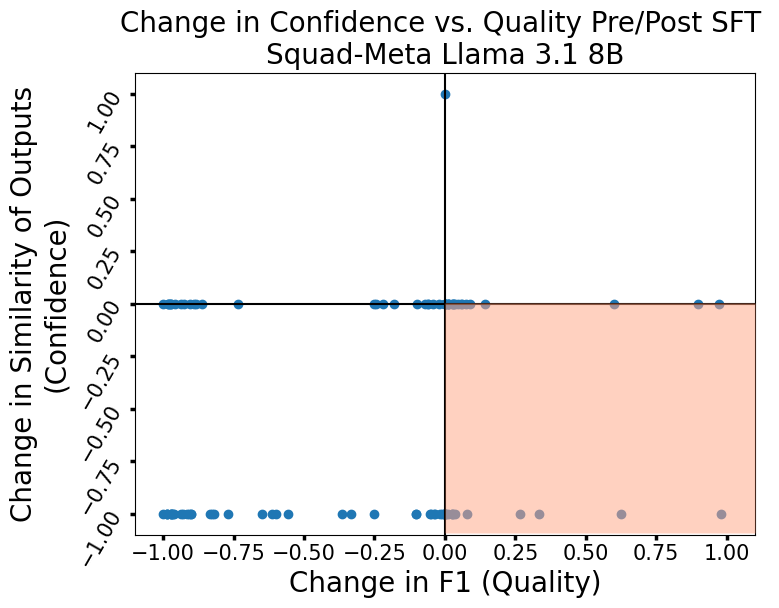}
  \caption{Dropout BLEU Var, BART (Left) and Llama 3.1 (Right) \label{fig:deltas_var}}
\end{subfigure}

\caption{Plots reveal that average log probabilities are more prone to relative overconfidence, while dropout BLEU variance is more prone to relative underconfidence after SFT; plotting change in F1 between epoch one and ten for SQUAD vs confidence score}
\label{fig:four_deltas}
\end{figure}

We compute the change in average log probs between the first epoch and post-SFT (See Table \ref{Table:calib_cls_log_prob}). We observe that although a bulk of the changes are concordant, there is a tendency for SFT to make BART, Llama-3.1, Gemma-2B, Gemma-9B, and QWen-14B relatively overconfident in samples. We visualize examples of relative overconfidence using the quadrant plots in Figure \ref{fig:deltas_logprobs}, which show that more samples fall in Quadrant II (Overconfident) compared to Quadrant IV (Underconfident).

\begin{table}[ht]\centering
\scriptsize
\centering
\begin{tabular}{llccc}\toprule

& \textbf{Dataset} & \textbf{Cncd} & \textbf{R. Over} & \textbf{R. Under} \\
\midrule

\multirow{3}{*}{\textbf{BART}}
&Eng-Afr &0.94 &0.06 &0.00 \\
&SQUAD &0.85 &0.11 &0.03 \\
&GSM8K &0.98 &0.02 &0.00 \\
\cmidrule(lr){2-5}
\multirow{3}{*}{\textbf{Flan-T5}} &Eng-Afr &0.17 &0.02 &0.80 \\
&SQUAD &0.82 &0.02 &0.16 \\
&GSM8K &0.99 &0.00 &0.01 \\
\cmidrule(lr){2-5}
\multirow{3}{*}{\textbf{Llama}} &Eng-Afr &0.55 &0.16 &0.29 \\
&SQUAD &0.37 &0.51 &0.12 \\
&GSM8K &0.90 &0.07 &0.03 \\
\cmidrule(lr){2-5}
\multirow{3}{*}{\textbf{Gemma-2B}} &Eng-Afr & 0.54 & 0.26 & 0.20 \\
&SQUAD & 0.63 & 0.27 & 0.10 \\
&GSM8K & 0.87 & 0.13 & 0.00 \\
\cmidrule(lr){2-5}
\multirow{3}{*}{\textbf{Gemma-9B}} &Eng-Afr & 0.69 & 0.28 & 0.03 \\
&SQUAD & 0.78 & 0.17 & 0.05 \\
&GSM8K & 0.92 & 0.08 & 0.00 \\
\cmidrule(lr){2-5}
\multirow{3}{*}{\textbf{QWen-14B}} &Eng-Afr & 0.56 & 0.30 & 0.14 \\
&SQUAD & 0.78 & 0.13 & 0.09 \\
&GSM8K & 0.89 & 0.04 & 0.07 \\
\bottomrule
\end{tabular}
\caption{Proportion of changes in test set samples that are concordant, relatively over/underconfident after SFT, using average token log probs as the measure of confidence; Reporting average proportions over 3 seeds}
\label{Table:calib_cls_log_prob}
\end{table}

To understand the changes in confidence score better, we plot each sample's average log probabilities by epoch. We observe that the distribution of the average log token probabilities shifts upwards with more epochs (See Figure \ref{fig:logprobs_by_epoch}).

Building on the pre/post-SFT analysis in Table \ref{Table:calib_cls_log_prob}, we classify epoch-level changes from epoch $t$ to epoch $t+1$ for $1 \leq t \leq 10$ in confidence score by epoch in Table \ref{Table:overconfidence_logprob_deepdive}. We find that (1) a large percentage of confidence scores are indeed increasing from epoch to epoch, but that among them, (2) quality is often decreasing, which corresponds to the relative overconfidence we observed in the pre/post SFT experiments. Our takeaway is that at the epoch-by-epoch level, fine-tuning pushes the confidence score (i.e. average log token probs) upward across \textit{most} outputs, regardless of the change in the output's quality, which may explain why the calibration between confidence and quality becomes worse.

\begin{table}[ht]\centering
\scriptsize
\centering
\begin{tabular}{lrrrrr}\toprule
\multicolumn{2}{r}{\textbf{Confidence increase?}} & \multicolumn{1}{c}{\textbf{No}} &\multicolumn{2}{c}{\textbf{Yes}} \\
\cmidrule(lr){4-5}
\multicolumn{2}{r}{\textbf{Quality improve?}} & & \textbf{No} & \textbf{Yes} \\
\midrule
\multirow{3}{*}{\textbf{BART}} &Eng-Afr &32.0 &26.3 &41.7 \\
&SQUAD &45.0 &45.5 &9.5  \\
&GSM8K &37.1 &61.4 &1.6 \\
\cmidrule{2-5}
\multirow{3}{*}{\textbf{Flan-T5}} &Eng-Afr &45.1 &27.5 &27.4 \\
&SQUAD &49.0 &47.2 &3.9 \\
&GSM8K &49.2 &49.9 &1.0 \\
\cmidrule{2-5}
\multirow{3}{*}{\textbf{Llama-3.1 8B}} &Eng-Afr &46.6 &23.9 &29.5 \\
&SQUAD &46.7 &44.9 &8.3 \\
&GSM8K &46.8 &35.4 &17.8 \\
\cmidrule{2-5}
\multirow{3}{*}{\textbf{Gemma-2B}} &Eng-Afr & 48.1 & 22.0 & 29.9 \\
&SQUAD & 43.9 & 37.3 & 18.8 \\
&GSM8K & 41.6 & 50.4 & 8.0 \\
\cmidrule{2-5}
\multirow{3}{*}{\textbf{Gemma-9B}} &Eng-Afr & 44.0 & 18.9 & 37.1 \\
&SQUAD & 48.8 & 35.8 & 15.4 \\
&GSM8K & 44.7 & 39.4 & 15.9 \\
\cmidrule{2-5}
\multirow{3}{*}{\textbf{QWen-14B}} &Eng-Afr & 45.2 & 21.0 & 33.8 \\
&SQUAD & 48.2 & 35.7 & 16.1 \\
&GSM8K & 47.5 & 40.4 & 12.1 \\
\bottomrule
\end{tabular}
\caption{Percentage of changes in test set samples between epochs $t$ and $t+1$ from $1 \leq t \leq 9$ \label{Table:overconfidence_logprob_deepdive} classified based on directionality of changes in confidence and quality; Reporting average percentages over 3 seeds}
\end{table}

\paragraph{Consistency-based methods are prone to relative underconfidence} We repeat the experiment using dropout variance with BLEU (Table \ref{Table:calib_cls_dropout_bleu}). Unlike the probability-based methods, the changes are either concordant or relatively underconfident. We show examples in Figure \ref{fig:deltas_var}, where we observe that more samples fall in Quadrant IV (Underconfident) than Quadrant II (Overconfident). When plotting the dropout BLEU variance values by epoch, we do not observe clear patterns (Fig \ref{fig:dropout_bleu_by_epoch}).

In summary, we find that after performing SFT, probability-based methods are more prone to relative overconfidence, whereas consistency-based methods are prone to relative underconfidence. Users may then choose which type of confidence metric to use, depending on which type of miscalibration is more acceptable.

\paragraph{\textcolor{red}{Case 1} relative miscalibration is the more common failure mode for probability-based scores} After studying the dynamics of confidence metrics pre/post SFT, we study how these changes result in miscalibration, to identify where interventions can be made. Using average log probs to measure confidence, Case 1 miscalibration happens more often than Case 2 in 10/18 dataset-model configurations (See Table \ref{Table:miscal_stats_logprobs}). When using variance in BLEU scores, Case 1 dominates in only 6/18 configurations (See Table \ref{Table:miscal_stats_bleu_var}), so the pattern holds for probability-based scores rather than across metric families.

\begin{table}[ht]
\scriptsize
\centering
\begin{tabular}{lrrrrr}
\toprule
\multicolumn{2}{r}{\textbf{Rel Qual.}} &\multicolumn{2}{c}{\textbf{Same}} &\multicolumn{2}{c}{\textbf{Flips}} \\
\cmidrule(lr){3-4} \cmidrule(lr){5-6}
\multicolumn{2}{r}{\textbf{Rel Conf.}} & Same & \textcolor{red}{\textbf{Flips}} &Flips & \textcolor{blue}{\textbf{Same}} \\
\midrule
\multirow{3}{*}{\textbf{BART}} &Eng-Afr & 0.70 & 0.12 & 0.11 & 0.06 \\
& SQUAD & 0.66 & 0.14 & 0.13 & 0.06 \\
& GSM8K & 0.79 & 0.16 & 0.03 & 0.01 \\
\cmidrule{2-6}
\multirow{3}{*}{\textbf{Flan-T5}} &Eng-Afr & 0.65 & 0.11 & 0.17 & 0.07 \\
& SQUAD & 0.74 & 0.18 & 0.06 & 0.02 \\
& GSM8K & 0.78 & 0.19 & 0.02 & 0.01 \\
\cmidrule{2-6}
\multirow{3}{*}{\textbf{Llama-3.1 8B}} &Eng-Afr & 0.39 & 0.12 & 0.38 & 0.11 \\
& SQUAD & 0.40 & 0.10 & 0.36 & 0.13 \\
& GSM8K & 0.66 & 0.19 & 0.11 & 0.05 \\
\cmidrule{2-6}
\multirow{3}{*}{\textbf{Gemma-2B}} &Eng-Afr & 0.38 & 0.11 & 0.39 & 0.12 \\
& SQUAD & 0.42 & 0.12 & 0.33 & 0.12 \\
& GSM8K & 0.61 & 0.17 & 0.15 & 0.06 \\
\cmidrule{2-6}
\multirow{3}{*}{\textbf{Gemma-9B}} &Eng-Afr & 0.47 & 0.00 & 0.27 & 0.26 \\
& SQUAD & 0.54 & 0.08 & 0.24 & 0.16 \\
& GSM8K & 0.69 & 0.02 & 0.19 & 0.16 \\
\cmidrule{2-6}
\multirow{3}{*}{\textbf{QWen-14B}} &Eng-Afr & 0.42 & 0.10 & 0.33 & 0.15 \\
& SQUAD & 0.57 & 0.06 & 0.20 & 0.15 \\
& GSM8K & 0.63 & 0.07 & 0.19 & 0.10 \\
\bottomrule
\end{tabular}
\caption{\textcolor{red}{\textbf{Case 1}} miscalibration happens more frequently than \textcolor{blue}{\textbf{Case 2}} in 10/18 configurations; Table shows proportion of datapoint pairs classified by change in relative quality and confidence between ep. $t$ and $t+1$; Results averaged across 10 epochs and 3 seeds, using average log probs as confidence scores \label{Table:miscal_stats_logprobs}}
\end{table}

\paragraph{\textcolor{red}{Case 1} miscalibration happens when the confidence score changes due to factors other than the output's quality} Probing more deeply, we find that in a considerable proportion of Case 1 miscalibrated pairs, the relative confidence scores incorrectly change, \textit{without} the quality of the actual prediction changing. In particular, we compute the proportion of Case 1 pairs, where either (1) the model increased its confidence in the worse sample, or (2) the model decreased its confidence in the better sample, without those samples' quality scores changing. We find that, when using BART or Flan-T5 for SQUAD and GSM8K, these cases comprise sometimes up to 99.9\% of miscalibrated cases (See Table \ref{Table:prop_miscal_no_change}).

Our takeaway here is that in many cases, miscalibration happens when the model's confidence scores change due to factors \textit{other} than the quality of the output itself. For example, higher model log probabilities in the output can indicate that the sample is closer to the training distribution of the model \citep{liu2020simpleprincipleduncertaintyestimation, vazhentsev-etal-2023-efficient}, which is not necessarily indicative of the correctness of the output. This is consistent with the mechanism described in prior work \citep{wang2024subjectiveuncertaintyquantificationcalibration}, where SFT sharpens the output distribution around a single target sequence, raising the probability on outputs that are in-distribution. We explore this in our experiments, by measuring the calibration between samples' confidence scores and their similarity to the training data. To compute similarity, we use the maximum cosine similarity between a sample and the samples in the training set. Indeed, for some datasets like SQUAD and GSM8K, there are small, but statistically significant positive calibrations between the model's confidence scores and the similarity of those samples to the training set (See Table \ref{Table:ood_test}). 

This suggests that it is not enough to use output probabilities and self-consistency methods to measure confidence, as these are affected by underlying factors beyond quality. Therefore, future work can propose methods that account for these interactions, hopefully to produce confidence scores that remain robust to SFT. Moreover, this shows how sensitive these metrics can be to various factors, highlighting how these confidence scores cannot simply be taken off-the-shelf without testing.

\section{Case Study: How does miscalibration affect downstream tasks?}
\label{section:case_study}

To illustrate the practical consequences of miscalibration, we evaluate confidence scores on a downstream application: using them to identify correct answers in a question-answering task. This represents a common real-world use case where confidence scores guide decision-making—for example, determining whether to present an answer to a user or flag it for review. Similar applications include hallucination detection and selective prediction, where UQ scores are routinely deployed \citep{manakul-etal-2023-selfcheckgpt, wang2023selfconsistencyimproveschainthought}.

\paragraph{Method} We use the TruthfulQA dataset \citep{lin2022truthfulqameasuringmodelsmimic} which consists of 817 question-answer pairs testing common misconceptions. We filter only to questions where the answers are entities (i.e. filtering out subjective questions and those where the answer is ``I am unable to answer'').

We divide the dataset into train-val-test splits, then perform SFT with early stopping. We obtain the model's answers on the test set, and get its confidence scores in those predictions. We use Claude 3.5 Sonnet to evaluate if the output matches the reference, since the answers are open-ended in ways that are hard to score with exact match or F1.

We rescale the confidence scores between 0\% and 100\% for better interpretability, so that users could be shown alerts like ``There is an estimated X\% chance the answer is right''. We use the min and max values of the score from the test set\footnote{In practice, the min/max values should be taken from the validation set, but because the questions on the test set are also in domain, we expect a similar distribution of scores.}.

\begin{table}[!ht]\centering
\resizebox{\columnwidth}{!}{
    \begin{tabular}{lrrrrrr}\toprule
    &\multicolumn{2}{c}{\textbf{BART}} &\multicolumn{2}{c}{\textbf{Flan-T5}} &\multicolumn{2}{c}{\textbf{Llama-3.1 8B}} \\
    \cmidrule(lr){2-3} \cmidrule(lr){4-5} \cmidrule(lr){6-7}
    & \textbf{1 Ep} & \textbf{Full} & \textbf{Pre} & \textbf{Full} & \textbf{Pre} & \textbf{Full} \\
    \midrule
    Tok Prob &0.563 &0.639 \textcolor{green}{$\uparrow$} &0.545 &0.610 \textcolor{green}{$\uparrow$} &0.532 &0.465 \textcolor{red}{$\downarrow$} \\
    Tok Ent &0.454 &0.330 \textcolor{red}{$\downarrow$} &0.492 &0.368 \textcolor{red}{$\downarrow$} &0.458 &0.512 \textcolor{green}{$\uparrow$} \\
    DO Ent &0.378 &0.463 \textcolor{green}{$\uparrow$} &0.068 &0.359 \textcolor{green}{$\uparrow$} &0.355 &0.453 \textcolor{green}{$\uparrow$} \\
    BS ImpWt &0.314 &0.437 \textcolor{green}{$\uparrow$} &0.511 &0.349 \textcolor{red}{$\downarrow$} &0.506 &0.492 \textcolor{red}{$\downarrow$} \\
    BS Rat &0.542 &0.560 \textcolor{green}{$\uparrow$} &0.609 &0.713 \textcolor{green}{$\uparrow$} &0.593 &0.409 \textcolor{red}{$\downarrow$} \\
    BS Sums &0.670 &0.542 \textcolor{red}{$\downarrow$} &0.486 &0.641 \textcolor{green}{$\uparrow$} &0.494 &0.508 \textcolor{green}{$\uparrow$} \\
    DO BLEU &0.411 &0.381 \textcolor{red}{$\downarrow$} &0.348 &0.533 \textcolor{green}{$\uparrow$} &0.554 &0.516 \textcolor{red}{$\downarrow$} \\
    DO KL &0.583 &0.489 \textcolor{red}{$\downarrow$} &0.185 &0.461 \textcolor{green}{$\uparrow$} &0.589 &0.470 \textcolor{red}{$\downarrow$} \\
    DO METr &0.580 &0.621 \textcolor{green}{$\uparrow$} &0.920 &0.512 \textcolor{red}{$\downarrow$} &0.465 &0.511 \textcolor{green}{$\uparrow$} \\
    CCo MSP &0.446 &0.226 \textcolor{red}{$\downarrow$} &0.606 &0.453 \textcolor{red}{$\downarrow$} &0.329 &0.492 \textcolor{green}{$\uparrow$} \\
    CCo MTE &0.429 &0.203 \textcolor{red}{$\downarrow$} &0.671 &0.492 \textcolor{red}{$\downarrow$} &0.335 &0.472 \textcolor{green}{$\uparrow$} \\
    CCo PPL &0.403 &0.207 \textcolor{red}{$\downarrow$} &0.606 &0.495 \textcolor{red}{$\downarrow$} &0.371 &0.484 \textcolor{green}{$\uparrow$} \\
    \midrule
    &\multicolumn{2}{c}{\textbf{Gemma 2-2B}} &\multicolumn{2}{c}{\textbf{Gemma 2-9B}} &\multicolumn{2}{c}{\textbf{QWen 2.5-14B}} \\
    \cmidrule(lr){2-3} \cmidrule(lr){4-5} \cmidrule(lr){6-7}
    & \textbf{1 Ep} & \textbf{Full} & \textbf{Pre} & \textbf{Full} & \textbf{Pre} & \textbf{Full} \\
    \midrule
    Tok Prob &0.455 &0.530 \textcolor{green}{$\uparrow$} &0.399 &0.517 \textcolor{green}{$\uparrow$} &0.687 &0.553 \textcolor{red}{$\downarrow$} \\
    Tok Ent &0.516 &0.449 \textcolor{red}{$\downarrow$} &0.593 &0.470 \textcolor{red}{$\downarrow$} &0.317 &0.461 \textcolor{green}{$\uparrow$} \\
    DO Ent &0.411 &0.401 \textcolor{red}{$\downarrow$} &0.456 &0.456 &0.332 &0.394 \textcolor{green}{$\uparrow$} \\
    BS ImpWt &0.569 &0.469 \textcolor{red}{$\downarrow$} &0.634 &0.522 \textcolor{red}{$\downarrow$} &0.312 &0.465 \textcolor{green}{$\uparrow$} \\
    BS Rat &0.618 &0.507 \textcolor{red}{$\downarrow$} &0.443 &0.482 \textcolor{green}{$\uparrow$} &0.603 &0.533 \textcolor{red}{$\downarrow$} \\
    BS Sums &0.418 &0.526 \textcolor{green}{$\uparrow$} &0.366 &0.480 \textcolor{green}{$\uparrow$} &0.688 &0.526 \textcolor{red}{$\downarrow$} \\
    DO BLEU &0.478 &0.470 \textcolor{red}{$\downarrow$} &0.518 &0.500 \textcolor{red}{$\downarrow$} &0.500 &0.517 \textcolor{green}{$\uparrow$} \\
    DO KL &0.510 &0.556 \textcolor{green}{$\uparrow$} &0.587 &0.568 \textcolor{red}{$\downarrow$} &0.608 &0.512 \textcolor{red}{$\downarrow$} \\
    DO METr &0.567 &0.440 \textcolor{red}{$\downarrow$} &0.509 &0.500 \textcolor{red}{$\downarrow$} &0.529 &0.452 \textcolor{red}{$\downarrow$} \\
    CCo MSP &0.492 &0.544 \textcolor{green}{$\uparrow$} &0.721 &0.448 \textcolor{red}{$\downarrow$} &0.334 &0.485 \textcolor{green}{$\uparrow$} \\
    CCo MTE &0.505 &0.537 \textcolor{green}{$\uparrow$} &0.705 &0.466 \textcolor{red}{$\downarrow$} &0.342 &0.473 \textcolor{green}{$\uparrow$} \\
    CCo PPL &0.522 &0.549 \textcolor{green}{$\uparrow$} &0.708 &0.473 \textcolor{red}{$\downarrow$} &0.337 &0.478 \textcolor{green}{$\uparrow$} \\
    \bottomrule
    \end{tabular}
    }
\caption{AUROC in detecting correct answers using various confidence metrics using the pre-trained and post-SFT model}\label{Table:case_study}
\end{table}

\paragraph{Results} We evaluate the AUROC in detecting the right answers. Pre/post SFT, calibration deteriorates after SFT in 35/72 cases (See Table \ref{Table:case_study}). We show an example in Figure \ref{Figure:sample}, where the model becomes rel. overconfident in a wrong answer post-SFT. Therefore, the usefulness or reliability of confidence scores can drop post-SFT if they are not designed so that their calibration is robust to SFT.

\section{Conclusion}

We investigate how supervised fine-tuning affects the calibration of confidence metrics for NLG tasks. Across 216 configurations, calibration degrades in 112 cases post-SFT and improves in 104, hence these metrics therefore cannot be deployed off-the-shelf without validation. We find that probability-based metrics exhibit relative overconfidence while consistency-based metrics show relative underconfidence, with miscalibration often occurring when confidence scores shift independently of output quality changes. Our downstream task evaluation shows performance degradation in 35/72 cases post-SFT. These findings reveal that confidence scores are influenced by factors beyond output quality, such as proximity to training data, motivating the need for calibration-robust UQ methods under model adaptation.

\section*{Limitations and Potential Risks}

While our study provides comprehensive evidence across 216 configurations spanning multiple models and tasks, several areas warrant further investigation. Our detailed mechanistic analyses in Section~\ref{section:overconfidence} focus on representative metrics (average log probabilities and BLEU variance) to deeply understand calibration dynamics; extending this analysis to all twelve metrics would strengthen these findings. Similarly, our investigation of miscalibration mechanisms emphasizes Case 1 (relative confidence flips); Case 2 merits additional study. We evaluate three diverse NLG tasks—translation, question answering, and mathematical reasoning—which span different generation characteristics; however, calibration sensitivity under SFT should be verified on additional task types such as summarization and dialogue.

Regarding evaluation methodology, PRR makes assumptions about the monotonic relationship between confidence and quality. While PRR is established for evaluating UQ in generation tasks \citep{malinin2021uncertaintyestimationautoregressivestructured, vashurin2025uncertaintyquantificationllmsminimum}, developing more sophisticated calibration metrics for NLG remains an important research direction. Our findings are robust to this limitation, as they demonstrate \textit{relative} changes in calibration rather than absolute values.

\bibliography{anthology,custom}

\appendix

\section{Confidence Metrics Equations}

\begin{equation}
    \label{Eq:Dropout_Entropy}
    \text{Conf} = \frac{1}{n_\text{dropout}} \sum_{i=1}^{n_\text{dropout}} \frac{1}{|\hat{y}^{(i)}|} \sum_{t=1}^{|\hat{y}^{(i)}|} \mathcal{H} \left(p(\hat{y}_t^{(i)} | \hat{y}_{<t}^{(i)}, x)\right)
\end{equation}

\begin{multline*}
    \mathcal{H}\left(p(\hat{y}_t^{(i)} | \hat{y}_{<t}^{(i)}, x)\right)= \\ - \sum_{j=1}^{|\mathcal{V}|} p(\hat{y}_{t,j}^{(i)} | \hat{y}_{<t}^{(i)}, x) \text{log} \left( p(\hat{y}_{t,j}^{(i)} | \hat{y}_{<t}^{(i)}, x)\right)
\end{multline*}

\begin{equation}
    \label{Eq:Top_K}
    \text{Conf} = -\sum_{i=1}^{10} \pi_i \left( \frac{1}{|\hat{y}^{(i)}|} \text{ln}(p(\hat{y}^{(i)})) \right)
\end{equation}

$$\pi_i = \frac{\text{exp}\left(\frac{1}{|\hat{y}^{(i)}|} \text{ln}(p(\hat{y}^{(i)}))\right)}{\sum_{j=1}^{10} \text{exp}\left(\frac{1}{|\hat{y}^{(j)}|} \text{ln}(p(\hat{y}^{(j)}))\right)}$$

$$\text{ln}(p(\hat{y}^{(i)})) = \sum_{t=1}^{|\hat{y}^{(i)}|} \text{ln}(p(\hat{y}_t^{(i)} | \hat{y}_{<t}^{(i)}, x))$$

\begin{equation}
    \label{Eq:METEOR_Var}
    \text{Conf} = \frac{\sum_{i=1}^{n_\text{Dropout}} \sum_{j=1}^{n_\text{Dropout}} \text{Meteor}(\hat{y}^{(i)}, \hat{y}^{(j)})}{N(N-1)} 
\end{equation}

\begin{equation}
    \label{Eq:BLEU_Var}
    \text{Conf} = \sum_{i=1}^{n_\text{Dropout}} \sum_{j=1}
^{n_\text{Dropout}} (1-\text{BLEU}(\hat{y}^{(i)}, \hat{y}^{(j)}))^2
\end{equation}

\begin{equation}
    \label{Eq:KL_Var}
    \text{Conf} = \sum_{i=1}^{n_\text{Dropout}} KL(p(\hat{y}^{(i)}|x), p_{\bar{y}})
\end{equation}

{
$$\bar{y}_\text{Prob}=\frac{1}{n_\text{Dropout}}\sum_{i=1}^{n_\text{Dropout}} p(\hat{y}^{(i)}|x)$$
}

Where $\hat{y}^{(i)}$ is the decoded sequence $i$ sampled by activating dropout, $\hat{y}_t^{(i)}$ is the $t$-th output token for sequence $i$, and $\hat{y}_{t,j}^{(i)}$ is the $j$-th vocabulary item at position $t$ for sequence $i$. When dropout is used, we sample $n_\text{Dropout}=3$ instances.

\paragraph{Interpreting the metrics} The metrics fall into two groups by sign. \textit{Higher-is-more-confident} metrics measure how concentrated the model's probability mass is on its chosen output: average token log probability, the beam-search importance weight (Eq \ref{Eq:Top_K}), the ratio between the top and $k$-th beam, and the sum of the top $k$ beams' probabilities. \textit{Higher-is-less-confident} metrics measure dispersion, either over the vocabulary (average token entropy, dropout entropy, Eq \ref{Eq:Dropout_Entropy}) or across dropout samples (BLEU and METEOR variance, Eqs \ref{Eq:BLEU_Var} and \ref{Eq:METEOR_Var}; KL divergence from the mean dropout distribution, Eq \ref{Eq:KL_Var}). CoCoA multiplies one metric of each kind. This sign convention explains the mirrored PRR values for paired metrics (e.g. Avg Tok Prob and Avg Tok Ent) in Table \ref{table:pre_last_calibration_delta}.

\section{Calibration Results}
\label{Appendix:calibration_details}
\begin{figure*}[htb]
\centering

\begin{subfigure}[htb]{\linewidth}
  \centering
  \includegraphics[width=\linewidth]{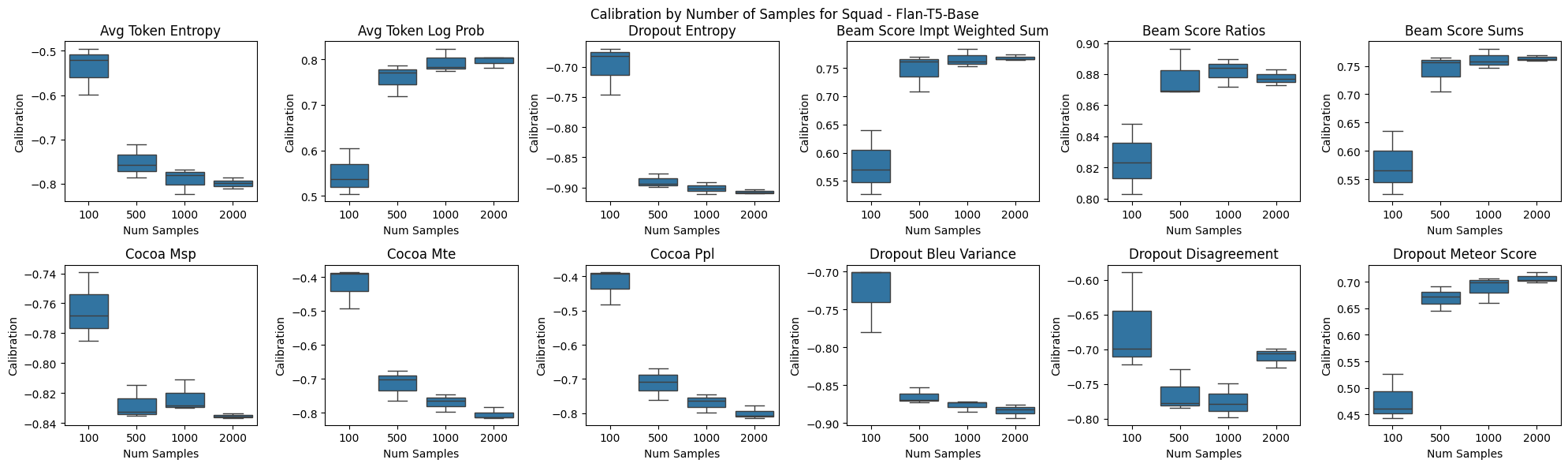}
  \caption{calibration of different confidence metrics after fine-tuning for 10 epochs.}
  \label{fig:calib_by_sample}
\end{subfigure}

\medskip

\begin{subfigure}[htb]{\linewidth}
  \centering
  \includegraphics[width=\linewidth]{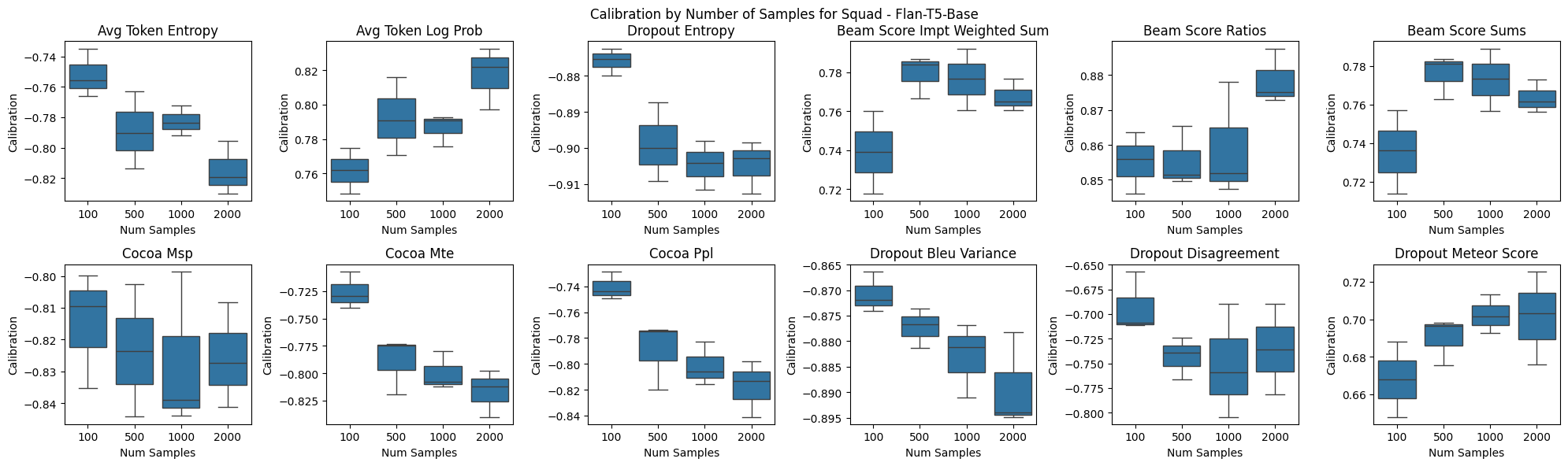}
  \caption{PRR of different confidence metrics after 2500 fine-tuning steps.}
  \label{fig:calib_by_ft_step}
\end{subfigure}

\caption{PRR of confidence metrics differs significantly depending on the number of fine-tuning samples. Plots shown for BART-Base fine-tuned on SQUAD for (top) 10 epochs and (bottom) 2500 steps using 3 seeds.}
\label{fig:calibration_comparison}
\end{figure*}

\begin{figure*}[htb]
  \centering
  \includegraphics[width=\linewidth]{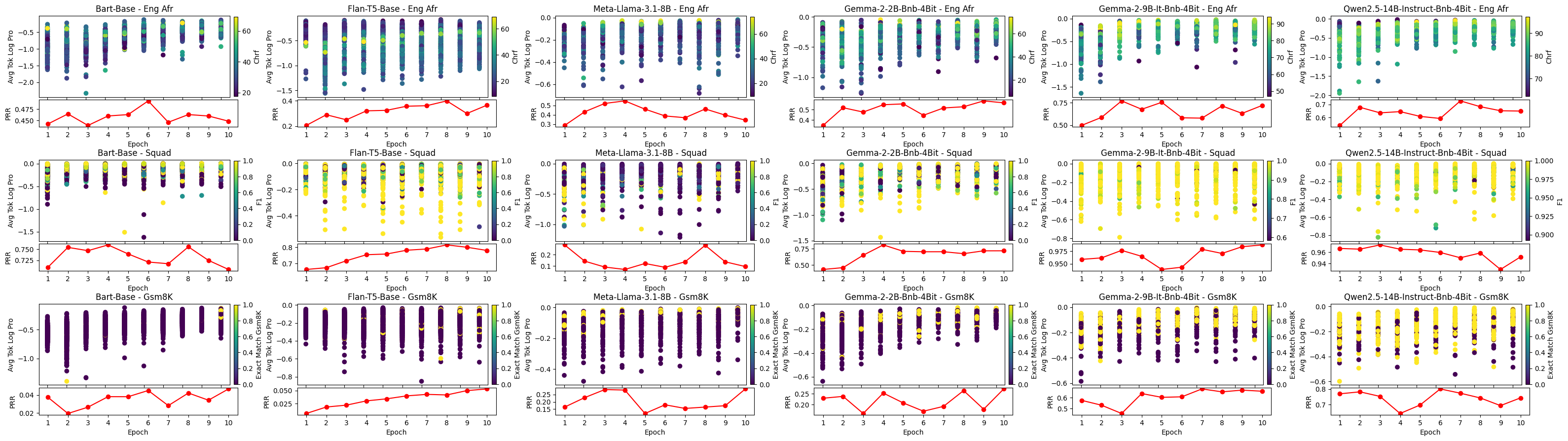}
  \caption{The average log probs across all test set samples generally increase for BART and Llama 3.1 8B (top), which may explain the rel. overconfidence in probability-based metrics, and fluctuations in calibration (bottom) \label{Figure:logprobs}}
\label{fig:logprobs_by_epoch}

\vspace{1em}
  \centering
  \includegraphics[width=\linewidth]{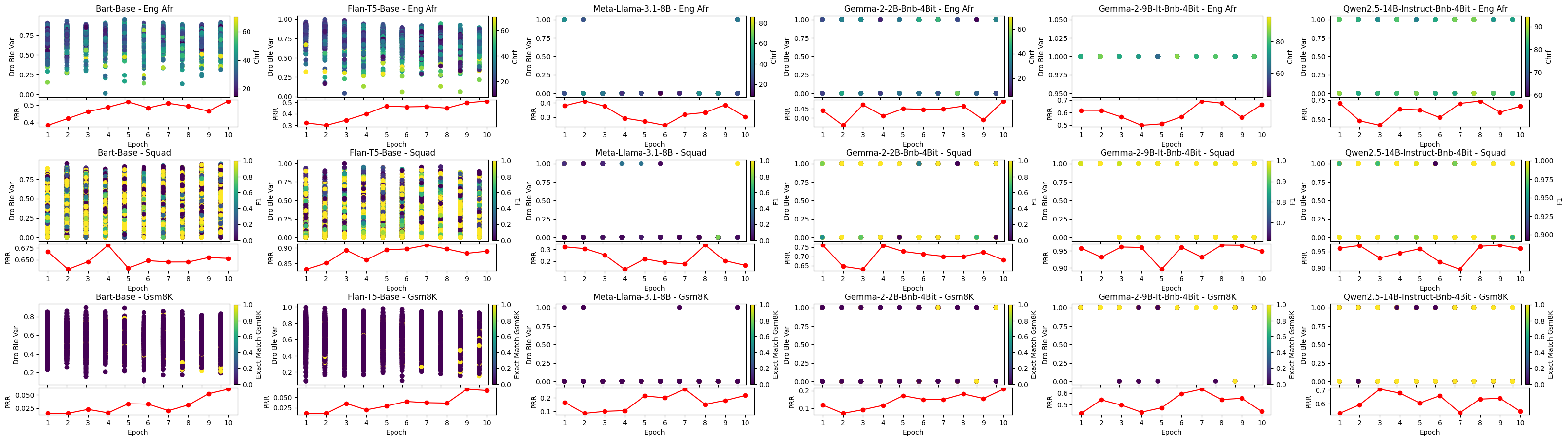}
  \caption{The dropout BLEU variance values generally do not change across epochs, which aligns with the fact that most samples remain calibrated post SFT.}
  \label{fig:dropout_bleu_by_epoch}
\end{figure*}

We report the maximum and minimum calibration by model and task, across the various confidence metrics in Tables \ref{table:first_last_calibration_delta}, \ref{Table:epoch_by_epoch_deltas}, \ref{Table:pre_first_last_gemma1}, \ref{Table:pre_first_last_gemma2}. We show the calibration by samples in Figure \ref{fig:calibration_comparison}. We show the plots of the average log probs and dropout BLEU variance values by epoch when fine-tuning with 2000 samples in Figure \ref{fig:logprobs_by_epoch} and Figure \ref{fig:dropout_bleu_by_epoch}.

\begin{table*}[ht]\centering

\resizebox{\linewidth}{!}{
    \begin{tabular}{lrrrrrrrrrr}\toprule
    &\multicolumn{3}{c}{\textbf{BART-Base}} &\multicolumn{3}{c}{\textbf{Flan-T5-Base}} &\multicolumn{3}{c}{\textbf{Llama 3.1-8B}} \\
    \cmidrule(lr){2-4} \cmidrule(lr){5-7} \cmidrule(lr){8-10}
    & \textbf{Eng-Afr} & \textbf{SQUAD} & \textbf{GSM8K} & \textbf{Eng-Afr} & \textbf{SQUAD} & \textbf{GSM8K} & \textbf{Eng-Afr} & \textbf{SQUAD} & \textbf{GSM8K} \\
    \midrule
    Avg Tok Prob & 0.131 (0.019) & 0.01 (0.011) & 0.016 (0.014) & 0.161 (0.021) & \textcolor{green}{0.131 (0.02)} & 0.019 (0.006) & 0.111 (0.033) & -0.115 (0.107) & 0.101 (0.053) \\
    Avg Tok Ent & \textcolor{red}{-0.132 (0.017)} & -0.011 (0.005) & -0.014 (0.013) & -0.159 (0.018) & \textcolor{red}{-0.139 (0.021)} & -0.018 (0.004) & -0.101 (0.033) & 0.116 (0.097) & -0.099 (0.056) \\
    DO Ent & -0.114 (0.026) & 0.002 (0.014) & -0.006 (0.007) & -0.194 (0.065) & -0.083 (0.023) & -0.02 (0.011) & -0.058 (0.027) & 0.17 (0.112) & -0.026 (0.027) \\
    BS Imp Wt & 0.124 (0.021) & -0.05 (0.033) & 0.007 (0.011) & 0.189 (0.031) & \textcolor{green}{0.095 (0.012)} & 0.013 (0.002) & 0.11 (0.026) & -0.099 (0.083) & 0.087 (0.053) \\
    BS Ratios & 0.11 (0.027) & 0.046 (0.013) & 0.016 (0.009) & 0.227 (0.054) & 0.011 (0.008) & 0.018 (0.012) & 0.03 (0.034) & -0.334 (0.102) & 0.103 (0.025) \\
    BS Sums & 0.124 (0.02) & -0.049 (0.036) & 0.007 (0.01) & 0.189 (0.031) & \textcolor{green}{0.096 (0.012)} & 0.013 (0.002) & 0.11 (0.026) & -0.096 (0.084) & 0.086 (0.055) \\
\cmidrule{2-10}
    DO Bleu Var & -0.108 (0.024) & -0.031 (0.014) & -0.01 (0.012) & -0.209 (0.045) & -0.063 (0.015) & -0.02 (0.003) & -0.02 (0.038) & 0.161 (0.088) & -0.029 (0.04) \\
    DO KL Div & -0.143 (0.007) & -0.047 (0.008) & -0.005 (0.01) & -0.207 (0.054) & -0.023 (0.06) & -0.011 (0.005) & -0.0 (0.039) & 0.239 (0.15) & -0.038 (0.035) \\
    DO Meteor Var & 0.131 (0.024) & 0.004 (0.042) & 0.008 (0.009) & 0.176 (0.04) & \textcolor{green}{0.164 (0.024)} & 0.016 (0.008) & 0.041 (0.056) & -0.173 (0.12) & 0.022 (0.085) \\
    CoCoA MSP & -0.125 (0.015) & -0.006 (0.022) & -0.01 (0.013) & -0.178 (0.041) & -0.015 (0.024) & -0.015 (0.007) & -0.028 (0.021) & 0.112 (0.167) & 0.007 (0.033) \\
    CoCoA MTE & -0.139 (0.035) & -0.021 (0.008) & -0.01 (0.008) & -0.143 (0.036) & -0.268 (0.058) & -0.015 (0.008) & -0.072 (0.072) & 0.16 (0.093) & -0.101 (0.059) \\
    CoCoA PPL & -0.135 (0.032) & -0.019 (0.009) & -0.011 (0.008) & -0.145 (0.034) & -0.265 (0.048) & -0.016 (0.009) & -0.081 (0.072) & 0.158 (0.103) & -0.103 (0.058) \\
    \bottomrule
    \end{tabular}
}

\resizebox{\linewidth}{!}{
    \begin{tabular}{lrrrrrrrrrr}\toprule
    &\multicolumn{3}{c}{\textbf{Gemma 2-2B}} &\multicolumn{3}{c}{\textbf{Gemma 2-9B}} &\multicolumn{3}{c}{\textbf{QWen 2.5-14B}} \\
    \cmidrule(lr){2-4} \cmidrule(lr){5-7} \cmidrule(lr){8-10}
    & \textbf{Eng-Afr} & \textbf{SQUAD} & \textbf{GSM8K} & \textbf{Eng-Afr} & \textbf{SQUAD} & \textbf{GSM8K} & \textbf{Eng-Afr} & \textbf{SQUAD} & \textbf{GSM8K} \\
    \midrule
    Avg Tok Prob & 0.147 (0.089) & 0.157 (0.196) & 0.002 (0.029) & 0.019 (0.159) & -0.007 (0.01) & 0.078 (0.059) & 0.023 (0.101) & -0.005 (0.01) & 0.007 (0.073) \\
    Avg Tok Ent & -0.149 (0.084) & -0.161 (0.208) & 0.015 (0.029) & -0.021 (0.139) & 0.006 (0.016) & -0.06 (0.061) & -0.024 (0.099) & 0.005 (0.012) & -0.011 (0.044) \\
    DO Ent & -0.081 (0.133) & -0.051 (0.18) & 0.015 (0.046) & 0.028 (0.23) & 0.021 (0.017) & -0.095 (0.022) & 0.061 (0.102) & 0.027 (0.009) & -0.051 (0.053) \\
    BS Imp Wt & 0.175 (0.086) & 0.033 (0.125) & -0.003 (0.016) & 0.007 (0.153) & -0.0 (0.024) & 0.092 (0.046) & 0.016 (0.087) & 0.004 (0.064) & 0.01 (0.063) \\
    BS Ratios & 0.067 (0.139) & 0.07 (0.118) & 0.048 (0.102) & -0.009 (0.253) & -0.002 (0.002) & -0.038 (0.017) & -0.065 (0.092) & -0.007 (0.006) & -0.0 (0.052) \\
    BS Sums & 0.175 (0.086) & 0.032 (0.127) & -0.005 (0.016) & 0.004 (0.156) & 0.002 (0.021) & 0.091 (0.048) & 0.015 (0.087) & 0.003 (0.067) & 0.01 (0.062) \\
\cmidrule{2-10}
    DO Bleu Var & -0.071 (0.087) & -0.052 (0.133) & 0.014 (0.035) & 0.047 (0.204) & 0.002 (0.002) & -0.036 (0.101) & 0.061 (0.081) & -0.005 (0.067) & 0.015 (0.09) \\
    DO KL Div & -0.061 (0.129) & -0.123 (0.146) & -0.013 (0.113) & 0.079 (0.206) & -0.0 (0.004) & -0.027 (0.134) & 0.079 (0.117) & 0.01 (0.026) & -0.049 (0.05) \\
    DO Meteor Var & 0.073 (0.122) & 0.103 (0.097) & 0.045 (0.027) & -0.046 (0.206) & 0.013 (0.004) & 0.04 (0.134) & -0.082 (0.108) & -0.014 (0.048) & 0.056 (0.032) \\
    CoCoA MSP & -0.003 (0.113) & -0.091 (0.093) & 0.008 (0.022) & 0.106 (0.238) & 0.01 (0.04) & -0.013 (0.027) & 0.068 (0.057) & -0.042 (0.068) & -0.096 (0.027) \\
    CoCoA MTE & -0.175 (0.09) & -0.119 (0.162) & -0.019 (0.062) & -0.006 (0.209) & 0.002 (0.021) & -0.082 (0.054) & 0.038 (0.101) & 0.014 (0.029) & -0.043 (0.051) \\
    CoCoA PPL & -0.172 (0.092) & -0.124 (0.165) & -0.026 (0.054) & -0.011 (0.213) & 0.004 (0.016) & -0.088 (0.045) & 0.029 (0.101) & 0.012 (0.019) & -0.04 (0.064) \\
    \bottomrule
    \end{tabular}
}

\caption{Difference in PRR after one epoch of SFT and after performing SFT with early stopping; Averaged across 3 seeds with st. dev in parentheses}\label{table:first_last_calibration_delta}
\end{table*}

\begin{table*}[ht]\centering

\resizebox{\linewidth}{!}{
    \begin{tabular}{lrrrrrrrrrr}\toprule
    &\multicolumn{3}{c}{\textbf{BART-Base}} &\multicolumn{3}{c}{\textbf{Flan-T5-Base}} &\multicolumn{3}{c}{\textbf{Llama 3.1-8B}} \\
    \cmidrule(lr){2-4} \cmidrule(lr){5-7} \cmidrule(lr){8-10}
    & \textbf{Eng-Afr} & \textbf{SQUAD} & \textbf{GSM8K} & \textbf{Eng-Afr} & \textbf{SQUAD} & \textbf{GSM8K} & \textbf{Eng-Afr} & \textbf{SQUAD} & \textbf{GSM8K} \\
    \midrule
    Avg Tok Prob & -0.080 (0.054) & -0.036 (0.013) & -0.022 (0.009) & -0.106 (0.030) & -0.029 (0.015) & -0.020 (0.007) & -0.178 (0.028) & -0.116 (0.029) & -0.171 (0.073) \\
    Avg Tok Ent & -0.079 (0.047) & -0.030 (0.007) & -0.021 (0.002) & -0.105 (0.030) & -0.074 (0.037) & -0.021 (0.009) & -0.120 (0.035) & -0.068 (0.064) & -0.161 (0.077) \\
    DO Ent & -0.071 (0.051) & -0.053 (0.014) & -0.016 (0.008) & -0.100 (0.020) & -0.041 (0.006) & -0.021 (0.008) & -0.136 (0.018) & -0.079 (0.053) & -0.145 (0.054) \\
    BS Imp Wt & -0.070 (0.041) & -0.049 (0.010) & -0.022 (0.006) & -0.115 (0.035) & -0.026 (0.004) & -0.014 (0.009) & -0.167 (0.033) & -0.121 (0.038) & -0.173 (0.076) \\
    BS Ratios & -0.070 (0.034) & -0.032 (0.012) & -0.021 (0.004) & -0.108 (0.064) & -0.162 (0.010) & -0.018 (0.007) & -0.104 (0.023) & -0.221 (0.051) & -0.197 (0.047) \\
    BS Sums & -0.070 (0.041) & -0.124 (0.041) & -0.021 (0.005) & -0.115 (0.035) & -0.107 (0.057) & -0.014 (0.009) & -0.166 (0.034) & -0.130 (0.090) & -0.173 (0.076) \\
    \cmidrule{2-10}
    DO BLEU Var & -0.067 (0.036) & -0.066 (0.015) & -0.026 (0.012) & -0.108 (0.046) & -0.034 (0.006) & -0.022 (0.011) & -0.092 (0.028) & -0.095 (0.061) & -0.120 (0.016) \\
    DO KL Div & -0.081 (0.033) & -0.053 (0.031) & -0.021 (0.012) & -0.090 (0.015) & -0.209 (0.086) & -0.026 (0.009) & -0.088 (0.016) & -0.267 (0.196) & -0.096 (0.015) \\
    DO Meteor Var & -0.072 (0.029) & -0.063 (0.010) & -0.019 (0.005) & -0.072 (0.004) & -0.024 (0.018) & -0.016 (0.005) & -0.120 (0.043) & -0.214 (0.133) & -0.119 (0.074) \\
    CoCoA MSP & -0.076 (0.039) & -0.056 (0.014) & -0.019 (0.008) & -0.088 (0.012) & -0.059 (0.012) & -0.016 (0.007) & -0.074 (0.013) & -0.251 (0.076) & -0.084 (0.039) \\
    CoCoA MTE & -0.078 (0.043) & -0.030 (0.007) & -0.020 (0.003) & -0.102 (0.025) & -0.127 (0.046) & -0.015 (0.004) & -0.128 (0.037) & -0.155 (0.047) & -0.180 (0.073) \\
    CoCoA PPL & -0.077 (0.046) & -0.032 (0.007) & -0.021 (0.004) & -0.104 (0.025) & -0.122 (0.036) & -0.015 (0.004) & -0.130 (0.038) & -0.167 (0.064) & -0.179 (0.062) \\
    \bottomrule
    \end{tabular}
}

\resizebox{\linewidth}{!}{
    \begin{tabular}{lrrrrrrrrrr}\toprule
    &\multicolumn{3}{c}{\textbf{Gemma 2-2B}} &\multicolumn{3}{c}{\textbf{Gemma 2-9B}} &\multicolumn{3}{c}{\textbf{QWen 2.5-14B}} \\
    \cmidrule(lr){2-4} \cmidrule(lr){5-7} \cmidrule(lr){8-10}
    & \textbf{Eng-Afr} & \textbf{SQUAD} & \textbf{GSM8K} & \textbf{Eng-Afr} & \textbf{SQUAD} & \textbf{GSM8K} & \textbf{Eng-Afr} & \textbf{SQUAD} & \textbf{GSM8K} \\
    \midrule
    Avg Tok Prob & -0.155 (0.016) & -0.200 (0.097) & -0.129 (0.031) & -0.141 (0.037) & -0.054 (0.018) & -0.158 (0.031) & -0.191 (0.071) & -0.031 (0.012) & -0.118 (0.007) \\
    Avg Tok Ent & -0.153 (0.010) & -0.201 (0.095) & -0.107 (0.030) & -0.144 (0.033) & -0.056 (0.023) & -0.149 (0.040) & -0.191 (0.070) & -0.043 (0.018) & -0.101 (0.024) \\
    DO Ent & -0.120 (0.016) & -0.179 (0.089) & -0.126 (0.036) & -0.178 (0.088) & -0.049 (0.011) & -0.187 (0.031) & -0.237 (0.105) & -0.055 (0.036) & -0.111 (0.033) \\
    BS Imp Wt & -0.159 (0.005) & -0.154 (0.073) & -0.121 (0.023) & -0.137 (0.046) & -0.090 (0.021) & -0.158 (0.033) & -0.197 (0.076) & -0.059 (0.033) & -0.108 (0.047) \\
    BS Ratios & -0.131 (0.013) & -0.076 (0.044) & -0.080 (0.048) & -0.223 (0.101) & -0.009 (0.005) & -0.159 (0.066) & -0.279 (0.124) & -0.018 (0.012) & -0.142 (0.014) \\
    BS Sums & -0.159 (0.005) & -0.152 (0.077) & -0.120 (0.023) & -0.137 (0.046) & -0.090 (0.020) & -0.158 (0.033) & -0.197 (0.077) & -0.062 (0.037) & -0.107 (0.046) \\
    \cmidrule{2-10}
    DO BLEU Var & -0.122 (0.010) & -0.156 (0.035) & -0.099 (0.046) & -0.189 (0.076) & -0.052 (0.011) & -0.131 (0.035) & -0.272 (0.140) & -0.036 (0.018) & -0.173 (0.067) \\
    DO KL Div & -0.110 (0.013) & -0.183 (0.051) & -0.096 (0.027) & -0.218 (0.047) & -0.061 (0.030) & -0.142 (0.054) & -0.287 (0.141) & -0.058 (0.011) & -0.148 (0.047) \\
    DO Meteor Var & -0.143 (0.059) & -0.142 (0.062) & -0.095 (0.031) & -0.189 (0.076) & -0.058 (0.009) & -0.166 (0.025) & -0.279 (0.127) & -0.044 (0.019) & -0.176 (0.046) \\
    CoCoA MSP & -0.110 (0.053) & -0.151 (0.100) & -0.083 (0.050) & -0.213 (0.103) & -0.070 (0.001) & -0.094 (0.059) & -0.290 (0.150) & -0.038 (0.009) & -0.132 (0.085) \\
    CoCoA MTE & -0.157 (0.008) & -0.194 (0.070) & -0.135 (0.078) & -0.156 (0.048) & -0.048 (0.023) & -0.192 (0.041) & -0.199 (0.077) & -0.057 (0.018) & -0.131 (0.026) \\
    CoCoA PPL & -0.157 (0.009) & -0.190 (0.072) & -0.132 (0.080) & -0.153 (0.045) & -0.048 (0.017) & -0.194 (0.042) & -0.196 (0.081) & -0.044 (0.015) & -0.130 (0.024) \\
    \bottomrule
    \end{tabular}
}

\caption{From epoch-to-epoch, there can be a considerable decline in calibration, reaching differences of up to 0.290 PRR points; Results averaged over 3 seeds with st. dev in parentheses \label{Table:epoch_by_epoch_deltas}}
\end{table*}

\begin{table*}[!h]\centering
\scriptsize
\resizebox{\linewidth}{!}{%
    \begin{tabular}{lrrrrrrrrr}\toprule
    & \multicolumn{3}{c}{\textbf{Eng-Afr}} & \multicolumn{3}{c}{\textbf{SQUAD}} & \multicolumn{3}{c}{\textbf{GSM8K}} \\
    & \textbf{No SFT} & \textbf{1 Ep} & \textbf{Full} & \textbf{No SFT} & \textbf{1 Ep} & \textbf{Full} & \textbf{No SFT} & \textbf{1 Ep} & \textbf{Full} \\
    \midrule
    Avg Tok Prob & 0.092 (0.014) & 0.347 (0.102) & 0.37 (0.047) & 0.064 (0.0) & 0.364 (0.045) & 0.345 (0.015) & 0.0 (0.0) & -0.016 (0.041) & 0.024 (0.037) \\
    Avg Tok Ent & 0.109 (0.015) & 0.321 (0.112) & 0.378 (0.051) & 0.048 (0.0) & 0.384 (0.05) & 0.379 (0.017) & 0.0 (0.0) & -0.007 (0.042) & 0.023 (0.029) \\
    DO Ent & 0.047 (0.099) & 0.125 (0.067) & 0.11 (0.094) & -0.24 (0.023) & -0.004 (0.074) & -0.059 (0.103) & 0.0 (0.0) & -0.022 (0.04) & -0.021 (0.036) \\
    BS Imp Wt & -0.235 (0.0) & -0.435 (0.091) & -0.448 (0.069) & -0.144 (0.0) & -0.127 (0.06) & 0.083 (0.03) & 0.0 (0.0) & -0.047 (0.012) & -0.001 (0.033) \\
    BS Ratios & -0.218 (0.0) & 0.03 (0.068) & -0.068 (0.046) & 0.239 (0.0) & 0.276 (0.017) & 0.369 (0.048) & 0.0 (0.0) & -0.019 (0.035) & 0.005 (0.046) \\
    BS Sums & 0.237 (0.0) & 0.435 (0.091) & 0.448 (0.067) & 0.094 (0.0) & -0.039 (0.046) & -0.238 (0.049) & 0.0 (0.0) & 0.048 (0.011) & 0.0 (0.033) \\
    \cmidrule{2-10}
    DO BLEU Var & 0.012 (0.125) & 0.385 (0.018) & 0.302 (0.037) & -0.047 (0.017) & 0.063 (0.014) & 0.099 (0.072) & 0.0 (0.0) & -0.008 (0.052) & 0.006 (0.052) \\
    DO KL Div & -0.026 (0.121) & -0.05 (0.053) & 0.042 (0.064) & 0.147 (0.02) & 0.245 (0.036) & 0.347 (0.022) & 0.0 (0.0) & 0.016 (0.042) & 0.013 (0.011) \\
    DO Meteor Var & -0.012 (0.094) & 0.397 (0.016) & 0.292 (0.07) & -0.017 (0.028) & 0.096 (0.009) & 0.132 (0.079) & 0.0 (0.0) & 0.009 (0.03) & 0.015 (0.013) \\
    CoCoA MSP & 0.128 (0.009) & 0.165 (0.12) & 0.191 (0.092) & 0.221 (0.0) & 0.414 (0.014) & 0.443 (0.034) & 0.0 (0.0) & 0.006 (0.029) & 0.002 (0.028) \\
    CoCoA MTE & 0.182 (0.005) & 0.225 (0.115) & 0.287 (0.072) & 0.06 (0.0) & 0.43 (0.016) & 0.464 (0.033) & 0.0 (0.0) & -0.005 (0.016) & 0.005 (0.022) \\
    CoCoA PPL & 0.167 (0.006) & 0.257 (0.116) & 0.295 (0.066) & 0.07 (0.0) & 0.42 (0.015) & 0.445 (0.031) & 0.0 (0.0) & -0.011 (0.017) & 0.006 (0.026) \\
    \bottomrule
    \end{tabular}
}

\resizebox{\linewidth}{!}{%
    \begin{tabular}{lrrrrrrrrr}\toprule
    & \multicolumn{3}{c}{\textbf{Eng-Afr}} & \multicolumn{3}{c}{\textbf{SQUAD}} & \multicolumn{3}{c}{\textbf{GSM8K}} \\
    & \textbf{No SFT} & \textbf{1 Ep} & \textbf{Full} & \textbf{No SFT} & \textbf{1 Ep} & \textbf{Full} & \textbf{No SFT} & \textbf{1 Ep} & \textbf{Full} \\
    \midrule
    Avg Tok Prob & -0.428 (0.0) & -0.172 (0.082) & -0.166 (0.249) & 0.036 (0.0) & -0.048 (0.042) & -0.038 (0.034) & 0.0 (0.0) & 0.012 (0.027) & 0.006 (0.027) \\
    Avg Tok Ent & -0.375 (0.0) & -0.105 (0.088) & -0.155 (0.259) & 0.027 (0.0) & -0.051 (0.037) & -0.034 (0.048) & 0.0 (0.0) & 0.013 (0.029) & 0.003 (0.017) \\
    DO Ent & -0.254 (0.044) & 0.116 (0.028) & 0.264 (0.11) & 0.313 (0.011) & 0.388 (0.011) & 0.397 (0.031) & 0.0 (0.0) & 0.036 (0.013) & 0.038 (0.043) \\
    BS Imp Wt & 0.557 (0.0) & 0.314 (0.04) & 0.131 (0.269) & -0.079 (0.0) & 0.015 (0.019) & 0.103 (0.015) & 0.0 (0.0) & 0.03 (0.027) & 0.023 (0.014) \\
    BS Ratios & 0.365 (0.0) & -0.05 (0.143) & 0.384 (0.184) & 0.725 (0.0) & 0.634 (0.045) & 0.374 (0.057) & 0.0 (0.0) & 0.014 (0.03) & 0.056 (0.052) \\
    BS Sums & -0.557 (0.0) & -0.314 (0.04) & -0.132 (0.269) & 0.053 (0.0) & -0.034 (0.018) & -0.124 (0.015) & 0.0 (0.0) & -0.03 (0.027) & -0.023 (0.014) \\
    \cmidrule{2-10}
    DO BLEU Var & 0.285 (0.031) & 0.21 (0.063) & 0.287 (0.033) & 0.359 (0.01) & 0.446 (0.028) & 0.397 (0.03) & 0.0 (0.0) & 0.01 (0.018) & 0.021 (0.038) \\
    DO KL Div & -0.08 (0.099) & 0.055 (0.094) & 0.09 (0.113) & 0.371 (0.016) & 0.324 (0.068) & 0.241 (0.075) & 0.0 (0.0) & 0.017 (0.015) & 0.012 (0.065) \\
    DO Meteor Var & 0.28 (0.051) & 0.244 (0.092) & 0.307 (0.086) & 0.354 (0.013) & 0.337 (0.024) & 0.269 (0.039) & 0.0 (0.0) & 0.007 (0.011) & 0.016 (0.016) \\
    CoCoA MSP & 0.346 (0.0) & 0.054 (0.024) & 0.168 (0.116) & -0.175 (0.0) & -0.379 (0.047) & -0.097 (0.064) & 0.0 (0.0) & 0.006 (0.035) & 0.006 (0.028) \\
    CoCoA MTE & 0.447 (0.0) & 0.036 (0.047) & -0.205 (0.228) & -0.178 (0.0) & -0.378 (0.048) & -0.083 (0.078) & 0.0 (0.0) & -0.004 (0.044) & -0.001 (0.019) \\
    CoCoA PPL & 0.346 (0.0) & -0.012 (0.024) & -0.218 (0.226) & -0.175 (0.0) & -0.379 (0.047) & -0.078 (0.076) & 0.0 (0.0) & -0.004 (0.045) & 0.001 (0.024) \\
    \bottomrule
    \end{tabular}
}

\resizebox{\linewidth}{!}{%
    \begin{tabular}{lrrrrrrrrr}\toprule
    & \multicolumn{3}{c}{\textbf{Eng-Afr}} & \multicolumn{3}{c}{\textbf{SQUAD}} & \multicolumn{3}{c}{\textbf{GSM8K}} \\
    & \textbf{No SFT} & \textbf{1 Ep} & \textbf{Full} & \textbf{No SFT} & \textbf{1 Ep} & \textbf{Full} & \textbf{No SFT} & \textbf{1 Ep} & \textbf{Full} \\
    \midrule
    Avg Tok Prob & 0.121 (0.039) & -0.012 (0.049) & 0.245 (0.099) & 0.068 (0.065) & -0.381 (0.041) & -0.275 (0.08) & 0.0 (0.0) & 0.123 (0.025) & 0.274 (0.103) \\
    Avg Tok Ent & 0.124 (0.039) & -0.006 (0.063) & 0.232 (0.085) & 0.051 (0.061) & -0.386 (0.035) & -0.279 (0.094) & 0.0 (0.0) & 0.114 (0.051) & 0.258 (0.091) \\
    DO Ent & 0.02 (0.137) & -0.02 (0.127) & 0.148 (0.22) & 0.131 (0.085) & -0.185 (0.101) & -0.016 (0.095) & 0.0 (0.0) & 0.009 (0.014) & -0.032 (0.068) \\
    BS Imp Wt & -0.126 (0.044) & 0.052 (0.048) & -0.237 (0.078) & -0.089 (0.067) & 0.419 (0.024) & 0.234 (0.122) & 0.0 (0.0) & -0.125 (0.028) & -0.264 (0.102) \\
    BS Ratios & 0.011 (0.098) & 0.173 (0.22) & 0.172 (0.178) & 0.111 (0.068) & 0.653 (0.078) & 0.401 (0.202) & 0.0 (0.0) & -0.013 (0.076) & 0.204 (0.097) \\
    BS Sums & 0.127 (0.044) & -0.051 (0.048) & 0.236 (0.078) & 0.09 (0.067) & -0.436 (0.017) & -0.238 (0.121) & 0.0 (0.0) & 0.126 (0.029) & 0.264 (0.102) \\
    \cmidrule{2-10}
    DO Bleu Var & -0.015 (0.094) & 0.026 (0.075) & 0.0 (0.0) & -0.13 (0.082) & -0.085 (0.088) & -0.07 (0.189) & 0.0 (0.0) & 0.019 (0.032) & 0.0 (0.0) \\
    DO KL Div & -0.114 (0.098) & -0.002 (0.042) & -0.116 (0.087) & -0.14 (0.068) & 0.056 (0.094) & 0.11 (0.115) & 0.0 (0.0) & 0.013 (0.054) & -0.018 (0.066) \\
    DO Meteor Var & -0.042 (0.088) & 0.007 (0.104) & -0.108 (0.179) & -0.022 (0.088) & -0.141 (0.056) & -0.174 (0.011) & 0.0 (0.0) & 0.032 (0.11) & 0.092 (0.08) \\
    CoCoA MSP & 0.207 (0.022) & 0.215 (0.143) & 0.201 (0.031) & 0.14 (0.121) & 0.335 (0.062) & -0.076 (0.165) & 0.0 (0.0) & 0.154 (0.014) & 0.275 (0.119) \\
    CoCoA MTE & 0.228 (0.033) & 0.032 (0.181) & 0.184 (0.054) & 0.059 (0.074) & -0.253 (0.031) & -0.196 (0.118) & 0.0 (0.0) & 0.146 (0.034) & 0.262 (0.109) \\
    CoCoA PPL & 0.207 (0.022) & 0.017 (0.179) & 0.201 (0.031) & 0.07 (0.079) & -0.26 (0.027) & -0.201 (0.1) & 0.0 (0.0) & 0.154 (0.014) & 0.275 (0.119) \\
    \bottomrule
    \end{tabular}
}

\caption{PRR of confidence metrics before SFT, after one epoch, and after full SFT with early stopping for BART, Flan-T5, Llama 3.1-8B, Gemma 2-2B, Gemma 2-9B, and QWen 2.5-14B}
\label{Table:pre_first_last_gemma1}
\end{table*}

\begin{table*}[!h]\centering
\scriptsize

\resizebox{\linewidth}{!}{%
    \begin{tabular}{lrrrrrrrrr}\toprule
    & \multicolumn{3}{c}{\textbf{Eng-Afr}} & \multicolumn{3}{c}{\textbf{SQUAD}} & \multicolumn{3}{c}{\textbf{GSM8K}} \\
    & \textbf{No SFT} & \textbf{1 Ep} & \textbf{Full} & \textbf{No SFT} & \textbf{1 Ep} & \textbf{Full} & \textbf{No SFT} & \textbf{1 Ep} & \textbf{Full} \\
    \midrule
    Avg Tok Prob & 0.0 (0.043) & -0.184 (0.253) & 0.327 (0.089) & 0.268 (0.048) & -0.018 (0.08) & 0.019 (0.18) & 0.0 (0.0) & 0.163 (0.077) & 0.155 (0.075) \\
    Avg Tok Ent & -0.043 (0.047) & -0.169 (0.242) & 0.314 (0.081) & 0.26 (0.04) & -0.006 (0.1) & 0.03 (0.187) & 0.0 (0.0) & 0.174 (0.07) & 0.135 (0.062) \\
    DO Ent & -0.143 (0.02) & 0.004 (0.032) & 0.049 (0.116) & 0.177 (0.056) & 0.006 (0.067) & -0.094 (0.081) & 0.0 (0.0) & 0.022 (0.072) & -0.017 (0.083) \\
    BS Imp Wt & -0.002 (0.043) & 0.173 (0.264) & -0.395 (0.079) & -0.263 (0.041) & -0.173 (0.111) & 0.116 (0.142) & 0.0 (0.0) & -0.218 (0.141) & -0.238 (0.081) \\
    BS Ratios & 0.098 (0.04) & 0.089 (0.216) & 0.018 (0.029) & 0.093 (0.107) & 0.339 (0.105) & 0.324 (0.075) & 0.0 (0.0) & 0.169 (0.042) & 0.255 (0.1) \\
    BS Sums & 0.003 (0.043) & -0.173 (0.265) & 0.394 (0.078) & 0.263 (0.041) & 0.129 (0.109) & -0.155 (0.139) & 0.0 (0.0) & 0.217 (0.139) & 0.232 (0.083) \\
    \cmidrule{2-10}
    DO Bleu Var & -0.068 (0.033) & -0.052 (0.088) & -0.02 (0.087) & -0.005 (0.009) & 0.027 (0.069) & -0.024 (0.074) & 0.0 (0.0) & -0.011 (0.066) & -0.044 (0.039) \\
    DO KL Div & 0.044 (0.045) & 0.109 (0.026) & 0.047 (0.064) & -0.111 (0.045) & -0.041 (0.114) & 0.037 (0.047) & 0.0 (0.0) & -0.021 (0.22) & -0.025 (0.119) \\
    DO Meteor Var & -0.063 (0.095) & -0.024 (0.086) & -0.025 (0.106) & -0.074 (0.059) & 0.026 (0.102) & -0.012 (0.073) & 0.0 (0.0) & -0.012 (0.088) & -0.068 (0.107) \\
    CoCoA MSP & 0.138 (0.063) & -0.12 (0.066) & 0.341 (0.039) & 0.297 (0.032) & 0.159 (0.06) & 0.072 (0.234) & 0.0 (0.0) & 0.183 (0.063) & 0.192 (0.091) \\
    CoCoA MTE & 0.143 (0.068) & -0.257 (0.2) & 0.34 (0.11) & 0.289 (0.023) & 0.095 (0.064) & 0.044 (0.195) & 0.0 (0.0) & 0.081 (0.1) & 0.091 (0.041) \\
    CoCoA PPL & 0.138 (0.063) & -0.262 (0.207) & 0.343 (0.109) & 0.297 (0.032) & 0.083 (0.046) & 0.033 (0.195) & 0.0 (0.0) & 0.079 (0.09) & 0.113 (0.063) \\
    \bottomrule
    \end{tabular}
}
\resizebox{\linewidth}{!}{%
    \begin{tabular}{lrrrrrrrrr}\toprule
    & \multicolumn{3}{c}{\textbf{Eng-Afr}} & \multicolumn{3}{c}{\textbf{SQUAD}} & \multicolumn{3}{c}{\textbf{GSM8K}} \\
    & \textbf{No SFT} & \textbf{1 Ep} & \textbf{Full} & \textbf{No SFT} & \textbf{1 Ep} & \textbf{Full} & \textbf{No SFT} & \textbf{1 Ep} & \textbf{Full} \\
    \midrule
    Avg Tok Prob & -0.703 (0.028) & -0.609 (0.094) & -0.630 (0.060) & -0.841 (0.014) & -0.959 (0.017) & -0.953 (0.004) & -0.500 (0.000) & -0.566 (0.028) & -0.626 (0.062) \\
    Avg Tok Ent & 0.717 (0.023) & 0.600 (0.114) & 0.619 (0.058) & 0.848 (0.013) & 0.960 (0.014) & 0.953 (0.006) & 0.500 (0.000) & 0.551 (0.026) & 0.629 (0.051) \\
    DO Ent & 0.704 (0.011) & 0.612 (0.156) & 0.584 (0.088) & 0.856 (0.016) & 0.973 (0.004) & 0.952 (0.014) & 0.500 (0.000) & 0.326 (0.077) & 0.421 (0.069) \\
    BS Imp Wt & 0.731 (0.007) & 0.610 (0.109) & 0.617 (0.049) & 0.867 (0.005) & 0.904 (0.034) & 0.904 (0.012) & 0.500 (0.000) & 0.550 (0.034) & 0.642 (0.072) \\
    BS Ratios & 0.749 (0.044) & 0.577 (0.182) & 0.568 (0.075) & 0.904 (0.008) & 0.992 (0.001) & 0.991 (0.002) & 0.500 (0.000) & 0.551 (0.023) & 0.514 (0.037) \\
    BS Sums & 0.730 (0.007) & 0.610 (0.110) & 0.614 (0.049) & 0.866 (0.005) & 0.900 (0.031) & 0.902 (0.011) & 0.500 (0.000) & 0.549 (0.034) & 0.640 (0.073) \\
    \cmidrule{2-10}
    DO BLEU Var & 0.706 (0.034) & 0.609 (0.148) & 0.545 (0.087) & 0.820 (0.016) & 0.957 (0.009) & 0.941 (0.012) & 0.500 (0.000) & 0.410 (0.072) & 0.424 (0.048) \\
    DO KL Div & 0.687 (0.036) & 0.623 (0.132) & 0.546 (0.080) & 0.801 (0.018) & 0.946 (0.020) & 0.945 (0.022) & 0.500 (0.000) & 0.403 (0.090) & 0.431 (0.043) \\
    DO Meteor Var & 0.706 (0.034) & 0.609 (0.148) & 0.543 (0.084) & 0.817 (0.016) & 0.954 (0.005) & 0.952 (0.005) & 0.500 (0.000) & 0.410 (0.072) & 0.481 (0.054) \\
    CoCoA MSP & 0.631 (0.016) & 0.574 (0.147) & 0.467 (0.099) & 0.748 (0.019) & 0.916 (0.006) & 0.907 (0.035) & 0.500 (0.000) & 0.244 (0.085) & 0.256 (0.067) \\
    CoCoA MTE & 0.706 (0.030) & 0.611 (0.145) & 0.617 (0.069) & 0.852 (0.018) & 0.958 (0.022) & 0.956 (0.001) & 0.500 (0.000) & 0.507 (0.043) & 0.589 (0.056) \\
    CoCoA PPL & 0.716 (0.029) & 0.605 (0.146) & 0.615 (0.073) & 0.860 (0.015) & 0.959 (0.017) & 0.954 (0.003) & 0.500 (0.000) & 0.505 (0.039) & 0.594 (0.040) \\
    \bottomrule
    \end{tabular}
}

\resizebox{\linewidth}{!}{%
    \begin{tabular}{lrrrrrrrrr}\toprule
    & \multicolumn{3}{c}{\textbf{Eng-Afr}} & \multicolumn{3}{c}{\textbf{SQUAD}} & \multicolumn{3}{c}{\textbf{GSM8K}} \\
    & \textbf{No SFT} & \textbf{1 Ep} & \textbf{Full} & \textbf{No SFT} & \textbf{1 Ep} & \textbf{Full} & \textbf{No SFT} & \textbf{1 Ep} & \textbf{Full} \\
    \midrule
    Avg Tok Prob & -0.494 (0.014) & -0.632 (0.056) & -0.657 (0.067) & -0.828 (0.010) & -0.962 (0.014) & -0.957 (0.007) & -0.500 (0.000) & -0.713 (0.015) & -0.724 (0.077) \\
    Avg Tok Ent & 0.494 (0.014) & 0.632 (0.055) & 0.655 (0.068) & 0.828 (0.010) & 0.958 (0.017) & 0.953 (0.010) & 0.500 (0.000) & 0.713 (0.034) & 0.720 (0.087) \\
    DO Ent & 0.420 (0.017) & 0.684 (0.006) & 0.624 (0.097) & 0.727 (0.027) & 0.970 (0.004) & 0.943 (0.005) & 0.500 (0.000) & 0.659 (0.018) & 0.710 (0.043) \\
    BS Imp Wt & 0.566 (0.006) & 0.636 (0.055) & 0.652 (0.058) & 0.857 (0.014) & 0.903 (0.030) & 0.907 (0.042) & 0.500 (0.000) & 0.712 (0.019) & 0.722 (0.078) \\
    BS Ratios & 0.569 (0.009) & 0.712 (0.033) & 0.647 (0.075) & 0.863 (0.018) & 0.993 (0.001) & 0.986 (0.006) & 0.500 (0.000) & 0.733 (0.059) & 0.732 (0.015) \\
    BS Sums & 0.564 (0.006) & 0.636 (0.055) & 0.651 (0.057) & 0.854 (0.014) & 0.899 (0.030) & 0.902 (0.044) & 0.500 (0.000) & 0.712 (0.020) & 0.722 (0.078) \\
    \cmidrule{2-10}
    DO BLEU Var & 0.439 (0.017) & 0.701 (0.047) & 0.629 (0.058) & 0.723 (0.025) & 0.955 (0.015) & 0.947 (0.028) & 0.500 (0.000) & 0.626 (0.092) & 0.619 (0.059) \\
    DO KL Div & 0.410 (0.026) & 0.688 (0.045) & 0.611 (0.099) & 0.767 (0.017) & 0.955 (0.005) & 0.946 (0.028) & 0.500 (0.000) & 0.620 (0.055) & 0.668 (0.028) \\
    DO Meteor Var & 0.482 (0.016) & 0.697 (0.046) & 0.624 (0.087) & 0.810 (0.029) & 0.964 (0.006) & 0.945 (0.036) & 0.500 (0.000) & 0.616 (0.069) & 0.699 (0.049) \\
    CoCoA MSP & 0.461 (0.008) & 0.615 (0.047) & 0.547 (0.075) & 0.654 (0.054) & 0.899 (0.046) & 0.941 (0.030) & 0.500 (0.000) & 0.468 (0.019) & 0.564 (0.036) \\
    CoCoA MTE & 0.377 (0.015) & 0.706 (0.050) & 0.668 (0.064) & 0.819 (0.010) & 0.951 (0.009) & 0.937 (0.031) & 0.500 (0.000) & 0.664 (0.019) & 0.707 (0.054) \\
    CoCoA PPL & 0.379 (0.013) & 0.696 (0.048) & 0.667 (0.066) & 0.821 (0.011) & 0.956 (0.011) & 0.944 (0.022) & 0.500 (0.000) & 0.667 (0.014) & 0.707 (0.063) \\
    \bottomrule
    \end{tabular}
}

\caption{PRR of confidence metrics before SFT, after one epoch, and after full SFT with early stopping for BART, Flan-T5, Llama 3.1-8B, Gemma 2-2B, Gemma 2-9B, and QWen 2.5-14B}
\label{Table:pre_first_last_gemma2}
\end{table*}

\section{Miscalibration Analyses}

We report details of the analyses referenced in Section \ref{section:overconfidence} in Tables \ref{Table:calib_cls_dropout_bleu}, \ref{Table:miscal_stats_bleu_var}, \ref{Table:prop_miscal_no_change}, and \ref{Table:ood_test}.

\begin{table}[ht]
\scriptsize
\centering
\begin{tabular}{llccc}\toprule
& \textbf{Dataset} & \textbf{Cncd} & \textbf{R. Over} & \textbf{R. Under} \\
\midrule
\multirow{3}{*}{\textbf{BART}} & Eng-Afr & 0.41 & 0.06 & 0.54 \\
& SQUAD & 0.80 & 0.08 & 0.12 \\
& GSM8K & 0.97 & 0.01 & 0.02\\
\cmidrule{2-5}
\multirow{3}{*}{\textbf{Flan-T5}} & Eng-Afr & 0.34 & 0.05 & 0.61 \\
& SQUAD & 0.88 & 0.04 & 0.08 \\
& GSM8K & 0.97 & 0.00 & 0.02 \\
\cmidrule{2-5}
\multirow{3}{*}{\textbf{Llama-3.1 8B}} & Eng-Afr & 0.94 & 0.00 & 0.06 \\
& SQUAD & 0.92 & 0.02 & 0.06 \\
& GSM8K & 1.00 & 0.00 & 0.00 \\
\cmidrule{2-5}
\multirow{3}{*}{\textbf{Gemma-2B}} & Eng-Afr & 0.73 & 0.01 & 0.26 \\
&SQUAD & 0.87 & 0.09 & 0.04 \\
&GSM8K & 0.92 & 0.01 & 0.07 \\
\cmidrule{2-5}
\multirow{3}{*}{\textbf{Gemma-9B}} & Eng-Afr & 1.00 & 0.00 & 0.00 \\
&SQUAD & 0.97 & 0.00 & 0.03 \\
&GSM8K & 0.99 & 0.00 & 0.01 \\
\cmidrule{2-5}
\multirow{3}{*}{\textbf{QWen-14B}} & Eng-Afr & 0.89 & 0.00 & 0.11 \\
&SQUAD & 0.92 & 0.07 & 0.01 \\
&GSM8K & 0.95 & 0.02 & 0.03 \\
\bottomrule
\end{tabular}
\caption{Proportion of changes in test set samples that are concordant, relatively over/underconfident after SFT, using dropout BLEU variance as the measure of confidence; Reporting average proportions over 3 seeds}
\label{Table:calib_cls_dropout_bleu}
\end{table}

\begin{table}[ht]
\centering
\scriptsize
\begin{tabular}{lrrrrr}
\toprule
\multicolumn{2}{r}{\textbf{Rel Qual.}} &\multicolumn{2}{c}{\textbf{Same}} &\multicolumn{2}{c}{\textbf{Flips}} \\
\cmidrule(lr){3-4} \cmidrule(lr){5-6}
\multicolumn{2}{r}{\textbf{Rel Conf.}} &Same & \textcolor{red}{\textbf{Flips}} &Flips & \textcolor{blue}{\textbf{Same}} \\
\midrule
\multirow{3}{*}{\textbf{BART}} &Eng-Afr & 0.70 & 0.13 & 0.12 & 0.06 \\
& SQUAD & 0.76 & 0.20 & 0.03 & 0.01 \\
& GSM8K & 0.63 & 0.18 & 0.14 & 0.05 \\
\cmidrule{2-6}
\multirow{3}{*}{\textbf{Flan-T5}} &Eng-Afr & 0.65 & 0.12 & 0.15 & 0.08 \\
& SQUAD & 0.79 & 0.18 & 0.02 & 0.01 \\
& GSM8K & 0.79 & 0.13 & 0.06 & 0.03 \\
\cmidrule{2-6}
\multirow{3}{*}{\textbf{Llama-3.1 8B}} &Eng-Afr & 0.48 & 0.05 & 0.24 & 0.22 \\
& SQUAD & 0.84 & 0.01 & 0.08 & 0.08 \\
& GSM8K & 0.48 & 0.03 & 0.27 & 0.22 \\
\cmidrule{2-6}
\multirow{3}{*}{\textbf{Gemma-2B}} &Eng-Afr & 0.41 & 0.08 & 0.34 & 0.17 \\
& SQUAD & 0.48 & 0.07 & 0.29 & 0.17 \\
& GSM8K & 0.70 & 0.08 & 0.13 & 0.08 \\
\cmidrule{2-6}
\multirow{3}{*}{\textbf{Gemma-9B}} &Eng-Afr & 0.47 & 0.00 & 0.27 & 0.26 \\
& SQUAD & 0.54 & 0.08 & 0.24 & 0.16 \\
& GSM8K & 0.69 & 0.02 & 0.19 & 0.16 \\
\cmidrule{2-6}
\multirow{3}{*}{\textbf{QWen-14B}} &Eng-Afr & 0.41 & 0.13 & 0.33 & 0.14 \\
& SQUAD & 0.57 & 0.06 & 0.20 & 0.15 \\
& GSM8K & 0.63 & 0.07 & 0.19 & 0.10 \\
\bottomrule
\end{tabular}
\caption{\textcolor{red}{\textbf{Case 1}} miscalibration happens more frequently than \textcolor{blue}{\textbf{Case 2}} in 6/18 configurations; Table shows proportion of datapoint pairs classified by change in relative quality and confidence between ep. $t$ and $t+1$; Results avg across 10 epochs and 3 seeds, using dropout BLEU variance \label{Table:miscal_stats_bleu_var}}
\end{table}

\begin{table*}[ht]\centering

\resizebox{\textwidth}{!}{
    \begin{tabular}{lrrrrrrrrrrrrr}\toprule
    &\multicolumn{4}{c}{\textbf{Eng-Afr}} &\multicolumn{4}{c}{\textbf{SQUAD}} &\multicolumn{4}{c}{\textbf{GSM8K}} \\
    \cmidrule(lr){2-5} \cmidrule(lr){6-9} \cmidrule(lr){10-13}
    \textbf{Ep} & \textbf{BART} & \textbf{T5} & \textbf{Llma} & \textbf{Gmma} & \textbf{BART} & \textbf{T5} & \textbf{Llma} & \textbf{Gmma} & \textbf{BART} & \textbf{T5} & \textbf{Llma} & \textbf{Gmma} \\\midrule
    2 &0.429 &0.182 &0.236 &0.013 &0.804 &0.875 &0.161 &0.394 &0.992 &0.989 &0.924 &0.924 \\
    3 &0.447 &0.277 &0.167 &0.262 &0.876 &0.908 &0.008 &0.430 &0.999 &0.990 &0.923 &0.941 \\
    4 &0.472 &0.426 &0.264 &0.135 &0.879 &0.935 &0.103 &0.586 &0.994 &0.989 &0.929 &0.912 \\
    5 &0.470 &0.245 &0.139 &0.226 &0.893 &0.926 &0.142 &0.501 &0.995 &0.994 &0.905 &0.931 \\
    6 &0.569 &0.429 &0.095 &0.148 &0.882 &0.961 &0.027 &0.673 &0.996 &0.993 &0.955 &0.927 \\
    7 &0.588 &0.428 &0.252 &0.153 &0.880 &0.948 &0.014 &0.399 &0.987 &0.992 &0.949 &0.898 \\
    8 &0.504 &0.270 &0.181 &0.113 &0.899 &0.958 &0.102 &0.507 &0.995 &0.989 &0.950 &0.921 \\
    9 &0.445 &0.409 &0.206 &0.165 &0.898 &0.976 &0.161 &0.595 &0.996 &0.988 &0.943 &0.955 \\
    10 &0.521 &0.359 &0.134 &0.234 &0.901 &0.968 &0.026 &0.616 &0.981 &0.990 &0.970 &0.929 \\
    \bottomrule
    \end{tabular}
}
\vspace{1em}

\resizebox{\textwidth}{!}{ 
    \begin{tabular}{lrrrrrrrrrrrrr}\toprule
    &\multicolumn{4}{c}{\textbf{Eng-Afr}} &\multicolumn{4}{c}{\textbf{SQUAD}} &\multicolumn{4}{c}{\textbf{GSM8K}} \\
    \cmidrule(lr){2-5} \cmidrule(lr){6-9} \cmidrule(lr){10-13}
    \textbf{Ep} & \textbf{BART} & \textbf{T5} & \textbf{Llma} & \textbf{Gmma} & \textbf{BART} & \textbf{T5} & \textbf{Llma} & \textbf{Gmma} & \textbf{BART} & \textbf{T5} & \textbf{Llma} & \textbf{Gmma} \\\midrule
    2 &0.470 &0.203 &0.126 &0.000 &0.744 &0.776 &0.130 &0.422 &0.994 &0.998 &0.000 &0.805 \\
    3 &0.356 &0.462 &0.000 &0.173 &0.862 &0.836 &0.000 &0.372 &0.994 &0.992 &0.000 &0.927 \\
    4 &0.445 &0.591 &0.000 &0.408 &0.878 &0.854 &0.000 &0.547 &0.990 &0.997 &0.000 &0.968 \\
    5 &0.642 &0.420 &0.000 &0.130 &0.912 &0.906 &0.000 &0.447 &0.985 &0.992 &0.000 &0.918 \\
    6 &0.477 &0.588 &0.000 &0.163 &0.883 &0.930 &0.000 &0.438 &0.989 &0.996 &1.000 &0.941 \\
    7 &0.522 &0.630 &0.000 &0.098 &0.879 &0.936 &0.000 &0.563 &0.992 &0.993 &1.000 &0.945 \\
    8 &0.563 &0.505 &0.000 &0.016 &0.893 &0.923 &0.000 &0.203 &0.990 &0.990 &0.000 &0.847 \\
    9 &0.450 &0.650 &0.000 &0.102 &0.904 &0.942 &0.000 &0.323 &0.985 &0.989 &0.000 &0.841 \\
    10 &0.506 &0.609 &0.000 &0.097 &0.886 &0.948 &0.000 &0.495 &0.988 &0.994 &0.000 &0.693 \\
    \bottomrule
    \end{tabular}
}

\caption{Proportion of Case 1 miscalibrated pairs where either the worse sample's output did not change, but the confidence score increased, or the better sample's output did not change, but the confidence score decreased, using average log probs (top) and variance in BLEU scores (bottom) as confidence scores \label{Table:prop_miscal_no_change}}
\end{table*}

\begin{table*}[ht]\centering
\begin{tabular}{lrrrrrrr}\toprule
&\multicolumn{2}{c}{\textbf{Eng-Afr}} &\multicolumn{2}{c}{\textbf{SQUAD}} &\multicolumn{2}{c}{\textbf{GSM8K}} \\
\cmidrule(lr){2-3} \cmidrule(lr){4-5} \cmidrule(lr){6-7}
\textbf{Epoch} & \textbf{BART} & \textbf{Flan-T5} & \textbf{BART} & \textbf{Flan-T5} & \textbf{BART} & \textbf{Flan-T5} \\
\midrule
1 &0.005 &-0.028 & *0.208 &0.043 & *0.121 &-0.086 \\
2 &0.057 &-0.127 & *0.165 & *0.154 & *0.206 &-0.082 \\
3 &-0.004 &-0.061 & *0.179 & *0.211 & *0.227 &-0.038 \\
4 &0.071 &-0.183 & *0.185 & *0.138 & *0.259 &-0.006 \\
5 &0.062 &-0.179 & *0.165 &0.079 & *0.177 &-0.053 \\
6 &0.107 &-0.112 &0.098 & *0.127 & *0.204 &-0.014 \\
7 &0.104 & *-0.226 &0.105 &0.094 & *0.323 &-0.021 \\
8 &0.100 &-0.215 & *0.122 &0.107 & *0.203 &-0.031 \\
9 &0.009 &-0.153 & *0.125 &0.110 & *0.252 &-0.013 \\
10 &0.013 &-0.051 & *0.160 & *0.111 & *0.188 &-0.017 \\
\bottomrule
\end{tabular}

\vspace{1em}

\begin{tabular}{lrrrrrrr}\toprule
&\multicolumn{2}{c}{\textbf{Eng-Afr}} &\multicolumn{2}{c}{\textbf{SQUAD}} &\multicolumn{2}{c}{\textbf{GSM8K}} \\
\cmidrule(lr){2-3} \cmidrule(lr){4-5} \cmidrule(lr){6-7}
\textbf{Epoch} & \textbf{BART} & \textbf{Flan-T5} & \textbf{BART} & \textbf{Flan-T5} & \textbf{BART} & \textbf{Flan-T5} \\
\midrule
1 &0.098 &-0.049 &-0.070 & *-0.166 &-0.099 &0.077 \\
2 &0.051 &0.025 &0.044 & *-0.168 & *-0.152 &0.035 \\
3 &0.053 &0.047 &-0.096 & *-0.205 & *-0.133 &0.067 \\
4 &0.140 &-0.020 &-0.083 & *-0.187 & *-0.157 &0.089 \\
5 &0.032 &-0.001 & *-0.120 & *-0.162 & *-0.163 &-0.012 \\
6 &0.173 &0.012 &-0.023 & *-0.177 & *-0.196 &0.097 \\
7 &0.173 &0.049 &-0.068 & *-0.160 & *-0.148 &0.059 \\
8 &0.066 &0.037 &-0.011 & *-0.181 & *-0.178 &0.030 \\
9 &0.110 &0.054 &-0.106 & *-0.180 & *-0.117 &-0.012 \\
10 &0.046 &0.076 &0.020 & *-0.167 & *-0.214 &0.051 \\
\bottomrule
\end{tabular}

\caption{PRR between confidence metric and the similarity of the sample to the training data, measured using the maximum cosine similarity between the last encoder hidden state embedding of the test set and a sample in the training set \label{Table:ood_test}}
\end{table*}

\section{Case Study Details}
\label{Appendix:Implementation}

\paragraph{Prompt} We use Anthropic Claude-3.5 Sonnet to rate whether or not the answer to the question is correct. We provide the original question, together with the reference answer in the dataset. We set the temperature to 0.

\texttt{Question: <question>\\
Correct Answer: <label>\\
Predicted Answer: <prediction>\\
Is the predicted answer correct, based on the correct answer?\\
Only return Yes or No}

\section{Dataset Details}
We use the NLLB dataset \cite{nllb-2022} under the ODC-By License, the FLORES Plus \cite{flores101} and SQUAD datasets \citep{rajpurkar-etal-2016-squad} under the CC BY-SA 4.0 License, and the GSM8K dataset \citep{cobbe2021gsm8k} under the MIT License, which allow the use of these datasets for research purposes. We scan the datasets to check that there is no malicious or harmful content in the translation pairs.

For translation, we use 10K samples from NLLB as the train set, an additional 100 samples from NLLB as the validation set, and 253 samples from FLORES as the test set. For math, we use 7K samples from the train set for training, an additional 100 samples from the train set as a validation set, and 1000 samples from the test set as the test. For SQUAD, we use 10K samples from the train set for training, and an additional 100 samples from the train set as a validation set. We use 1000 samples from the test set as the test.

\section{Computational Details}

Unless otherwise specified, we use a batch size of 8 and constant learning rate of 5e-5. We train models for a maximum of 200 epochs, but employ early stopping with a patience of 2 epochs; training is stopped once the validation metric does not increase on a validation set of 100 samples. We perform all fine-tuning and inference using one RTX 8000 GPU.

\end{document}